\documentclass[10pt]{article}
\usepackage[letterpaper,margin=0.85in]{geometry}
\usepackage{graphicx,booktabs,microtype,xcolor,amsmath,siunitx,caption,array,longtable,float}
\usepackage[hidelinks]{hyperref}

\newcommand{\BrainTestN}{433}
\newcommand{\BrainModelParamsM}{3.27}
\newcommand{\BrainMAE}{5.98}
\newcommand{\BrainMAECILow}{5.54}
\newcommand{\BrainMAECIHigh}{6.46}
\newcommand{\BrainRMSE}{7.67}
\newcommand{\BrainRMSECILow}{7.11}
\newcommand{\BrainRMSECIHigh}{8.30}
\newcommand{\BrainRtwo}{0.868}
\newcommand{\BrainRtwoCILow}{0.839}
\newcommand{\BrainRtwoCIHigh}{0.890}
\newcommand{\BrainPearson}{0.931}
\newcommand{\BrainPearsonCILow}{0.917}
\newcommand{\BrainPearsonCIHigh}{0.943}
\newcommand{\BrainSpearman}{0.931}
\newcommand{\BrainMedianAE}{4.94}
\newcommand{\BrainMeanDelta}{0.02}
\newcommand{\BrainDeltaSD}{7.68}
\newcommand{\BrainSagMAE}{7.47}
\newcommand{\BrainSagRMSE}{9.54}
\newcommand{\BrainCorMAE}{7.44}
\newcommand{\BrainCorRMSE}{9.43}
\newcommand{\BrainAxMAE}{7.49}
\newcommand{\BrainAxRMSE}{9.65}
\newcommand{\BrainFemaleN}{234}
\newcommand{\BrainFemaleMAE}{5.85}
\newcommand{\BrainMaleN}{199}
\newcommand{\BrainMaleMAE}{6.14}

\newcommand{\BrainResultsParagraph}{%
The three-axis fused estimate achieved a mean absolute error (MAE) of \BrainMAE{} years (95\% bootstrap CI, \BrainMAECILow--\BrainMAECIHigh), root-mean-square error (RMSE) of \BrainRMSE{} years (\BrainRMSECILow--\BrainRMSECIHigh), $R^2=\BrainRtwo$ (\BrainRtwoCILow--\BrainRtwoCIHigh), Pearson $r=\BrainPearson$ (\BrainPearsonCILow--\BrainPearsonCIHigh) and Spearman $\rho=\BrainSpearman$. The median absolute error was \BrainMedianAE{} years, and the mean brain-age delta was \BrainMeanDelta{} years with a standard deviation of \BrainDeltaSD{} years. The sagittal, coronal and axial axis-specific MAEs were \BrainSagMAE{}, \BrainCorMAE{} and \BrainAxMAE{} years, respectively. Descriptive sex-stratified MAEs were \BrainFemaleMAE{} years in \BrainFemaleN{} female scans and \BrainMaleMAE{} years in \BrainMaleN{} male scans. The cross-view summary is shown in Fig.~\ref{fig:brainage-overview}; agreement and age-stratified displays are shown in Fig.~\ref{fig:brainage} and Table~\ref{tab:brainage}.%
}

\newenvironment{summaryparagraph}{\begin{quote}\bfseries}{\end{quote}}

\title{From the Drosophila Visual Connectome to General-Purpose Computer Vision}
\author{Zongyu Li$^{1\dagger}$, Akito Yamauchi$^{2\dagger}$, Huaizhi Liu$^{1}$, Vishwanatha Rao$^{4,5}$, Jia Guo$^{1,3*}$,\\
for the Frontotemporal Lobar Degeneration Neuroimaging Initiative$^{\ddagger}$ and\\
for the Alzheimer's Disease Neuroimaging Initiative$^{**}$\\[0.45em]
\small $^{1}$Department of Biomedical Engineering, Columbia University, New York, NY, USA\\
\small $^{2}$Department of Electrical Engineering, Columbia University, New York, NY, USA\\
\small $^{3}$Department of Psychiatry, Columbia University, New York, NY, USA\\
\small $^{4}$Perelman School of Medicine, University of Pennsylvania, Philadelphia, PA, USA\\
\small $^{5}$Department of Bioengineering, School of Engineering and Applied Science,\\
\small University of Pennsylvania, Philadelphia, PA, USA\\
\small $^{\dagger}$These authors contributed equally to this work.\\
\small $^{*}$Corresponding author: \href{mailto:jg3400@columbia.edu}{jg3400@columbia.edu}}
\date{}
\hypersetup{
  pdftitle={From the Drosophila Visual Connectome to General-Purpose Computer Vision},
  pdfauthor={Zongyu Li; Akito Yamauchi; Huaizhi Liu; Vishwanatha Rao; Jia Guo; for the Frontotemporal Lobar Degeneration Neuroimaging Initiative; for the Alzheimer's Disease Neuroimaging Initiative},
  pdfsubject={Connectome-informed FlyVision models across natural and biomedical vision},
  pdfkeywords={ConnectomeX, FlyVision, Drosophila, ImageNet, biomedical imaging, brain age}
}
\begin{document}
\maketitle
\begin{center}
\parbox{0.93\textwidth}{\centering\footnotesize $^{\ddagger}$Data used in preparation of this article were obtained from the Frontotemporal Lobar Degeneration Neuroimaging Initiative (FTLDNI/NIFD) database. Investigators within FTLDNI/NIFD contributed to the design and implementation of the source initiative and/or provided data but did not participate in the ConnectomeX analysis or writing unless otherwise listed; the contributing investigator group is listed in the Acknowledgements.}\par
\vspace{0.35em}
\parbox{0.93\textwidth}{\centering\footnotesize $^{**}$Data used in preparation of this article were obtained from the Alzheimer's Disease Neuroimaging Initiative (ADNI) database (\href{https://adni.loni.usc.edu}{adni.loni.usc.edu}). The investigators within ADNI contributed to the design and implementation of ADNI and/or provided data but did not participate in the analysis or writing of this report. A complete listing of ADNI investigators is available from the \href{http://adni.loni.usc.edu/wp-content/uploads/how_to_apply/ADNI_Acknowledgement_List.pdf}{ADNI investigator list}.}
\end{center}

\begin{summaryparagraph}
Biological connectomes encode structured solutions to visual computation, raising the possibility that circuit organization can provide reusable inductive biases for artificial vision. Here we develop ConnectomeX around FlyVision, a trainable architecture that preserves parallel ON/OFF processing, recurrent computation and population-level graph interaction while scaling model capacity across tasks. FlyVision reached 99.34\% accuracy on MNIST with 80,608 trainable parameters and 78.03\% on CIFAR-10 with 81,408 parameters, compared with 99.22\% and 83.44\% for scratch-trained ResNet18 with 11.2 million parameters; transfer direction and magnitude varied across architectures. On ImageNet-1K validation, FlyVision Base and Large reached 60.79\% and 66.25\% top-1 accuracy with 1.8 and 3.7 million parameters, and a Large local-k7 model with a learned low-frequency branch reached 66.53\%, compared with 69.25\% for ResNet18 with 11.7 million parameters. On an exact-content-audited 22-class skin-disease split, FlyVision Large achieved 63.78\% accuracy, 60.23\% macro-F1 and 95.28\% macro-AUROC with 2.99 million parameters, compared with 66.19\%, 61.96\% and 95.96\% for ResNet18 with 11.19 million. In four-class chest radiography, ImageNet-pretrained FlyVision Base and Large reached 92.60\% and 92.76\% accuracy with 1.33 and 2.97 million parameters, respectively, compared with 91.56\% for ImageNet-pretrained ResNet18 with 11.18 million; hierarchical analysis localized most residual error to normal-versus-disease routing. The BrainAGE arm extends FlyVision to volumetric structural MRI: a shared ImageNet-pretrained FlyVision Large encoder processes 24 sagittal, coronal and axial slices from each T1-weighted (T1w) MRI volume and combines slice-level age predictions by confidence-modulated Gaussian voting. On the held-out T1w MRI test set, three-axis fusion achieved an MAE of \BrainMAE{} years and $R^2=\BrainRtwo$. Across the 224$\times$224 classification tasks, the best FlyVision configuration was within three percentage points of ResNet18 on ImageNet-1K and skin-disease classification and exceeded it on chest radiography with parameter counts approximately three- to fourfold lower. Together, these results show that a conserved connectome-informed computation can scale from compact recognition to large-scale natural and biomedical vision, with BrainAGE extending the framework to scan-level analysis of three-dimensional T1w MRI through multi-view aggregation.
\end{summaryparagraph}

Architectural priors specify the computations that a neural network can express efficiently before training begins. Convolution imposes locality and translation structure, residual connections alter information and gradient flow, and vision transformers provide global token interactions \cite{lecun1998,krizhevsky2012,he2016,dosovitskiy2021}. Biological visual systems embody a complementary design space: parallel channels, recurrent loops, heterogeneous cell populations and sparse structured connectivity assembled under simultaneous constraints on wiring, energy, latency and robustness. Modern connectomics converts these principles from qualitative biological motifs into experimentally measured circuit diagrams, creating the opportunity to ask whether biological wiring can inform the design of artificial visual systems.

Drosophila melanogaster provides a particularly tractable substrate for this translation. The FlyWire adult-brain reconstruction contains 139,255 neurons and approximately 54.5 million synapses \cite{dorkenwald2024}; systematic reconstruction of the optic lobe resolves approximately 38,500 intrinsic neurons into 227 cell types and their connectivity rules \cite{matsliah2024}; and whole-brain annotation extends cell-type and connectivity information across the central brain and optic lobes \cite{schlegel2024}. Connectome-constrained deep mechanistic models have further demonstrated that measured fly visual connectivity can support task-optimized networks that reproduce neural responses in the motion pathway \cite{lappalainen2024}. Biological connectomes have also been explored as direct representations for artificial-network architecture and biofidelic computation \cite{schmidgall2022,yu2026}. These advances motivate a complementary engineering question: whether abstractions of the same circuit organization can remain computationally useful outside the biological tasks and stimulus distributions from which the wiring was measured.

\begin{figure}[!htbp]\centering
\includegraphics[width=0.98\textwidth]{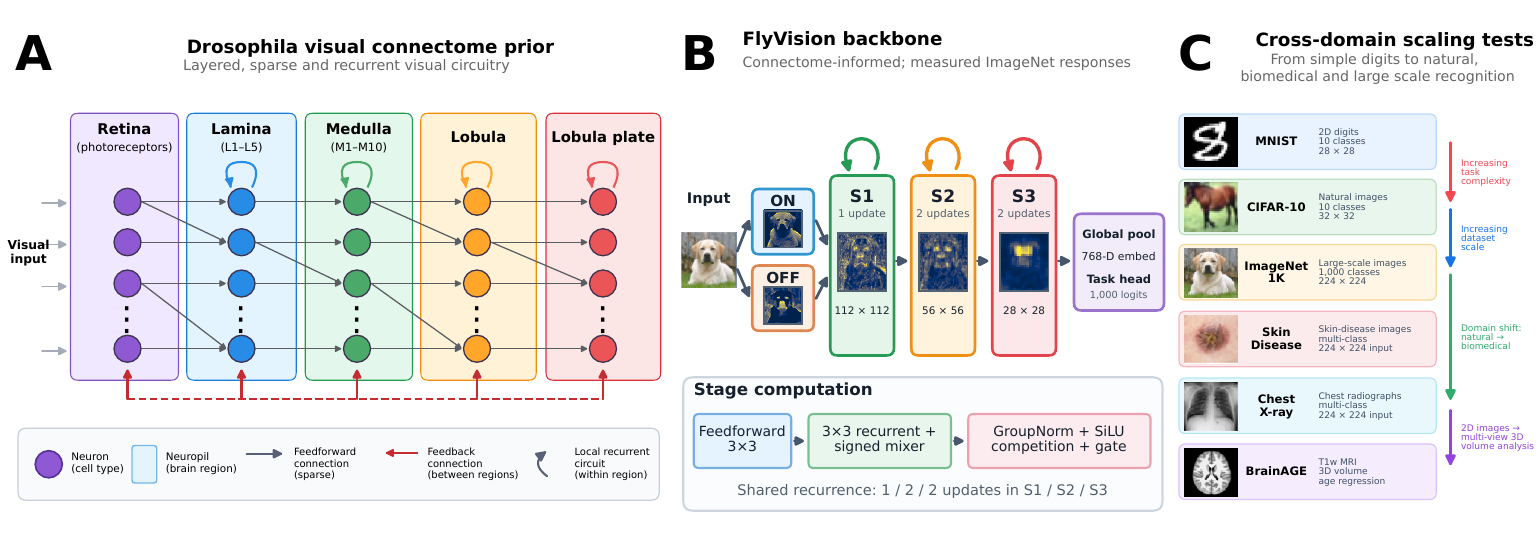}
\caption{Connectome-informed visual computation and cross-domain scaling. (A) Population-level schematic of the \textit{Drosophila} visual-connectome prior. Representative neuronal populations are arranged across the retina, lamina, medulla, lobula and lobula plate. Solid arrows indicate sparse feed-forward projections between successive neuropils, curved arrows denote local recurrent processing, and red dashed arrows summarize inter-region feedback. (B) ImageNet-trained FlyVision Large backbone. The input is the dog image already shown in panel C. ON and OFF stem maps and stages S1--S3 show mean absolute activations across all channels from a forward pass through the trained global-surround checkpoint. Each map uses its own first--99th percentile display range; color intensity is not comparable across maps. Each stage has a feed-forward $3\times3$ convolution and recurrent updates combining a $3\times3$ convolution with a signed four-population mixer, followed by GroupNorm, SiLU, lateral competition and a gated state update. S1/S2/S3 use one/two/two updates with shared weights. Global pooling, a 768-dimensional embedding and the ImageNet head produce 1,000 logits for this example. (C) Cross-domain scaling tests spanning MNIST, CIFAR-10, ImageNet-1K, skin-disease imaging, chest radiography and volumetric BrainAGE regression. BrainAGE is represented as multi-view analysis of a three-dimensional T1-weighted MRI volume: the model processes 24 slices across sagittal, coronal and axial planes and fuses them to a scan-level age estimate.}\label{fig:connectome}
\end{figure}

\begin{figure}[!htbp]\centering
\includegraphics[width=0.99\textwidth]{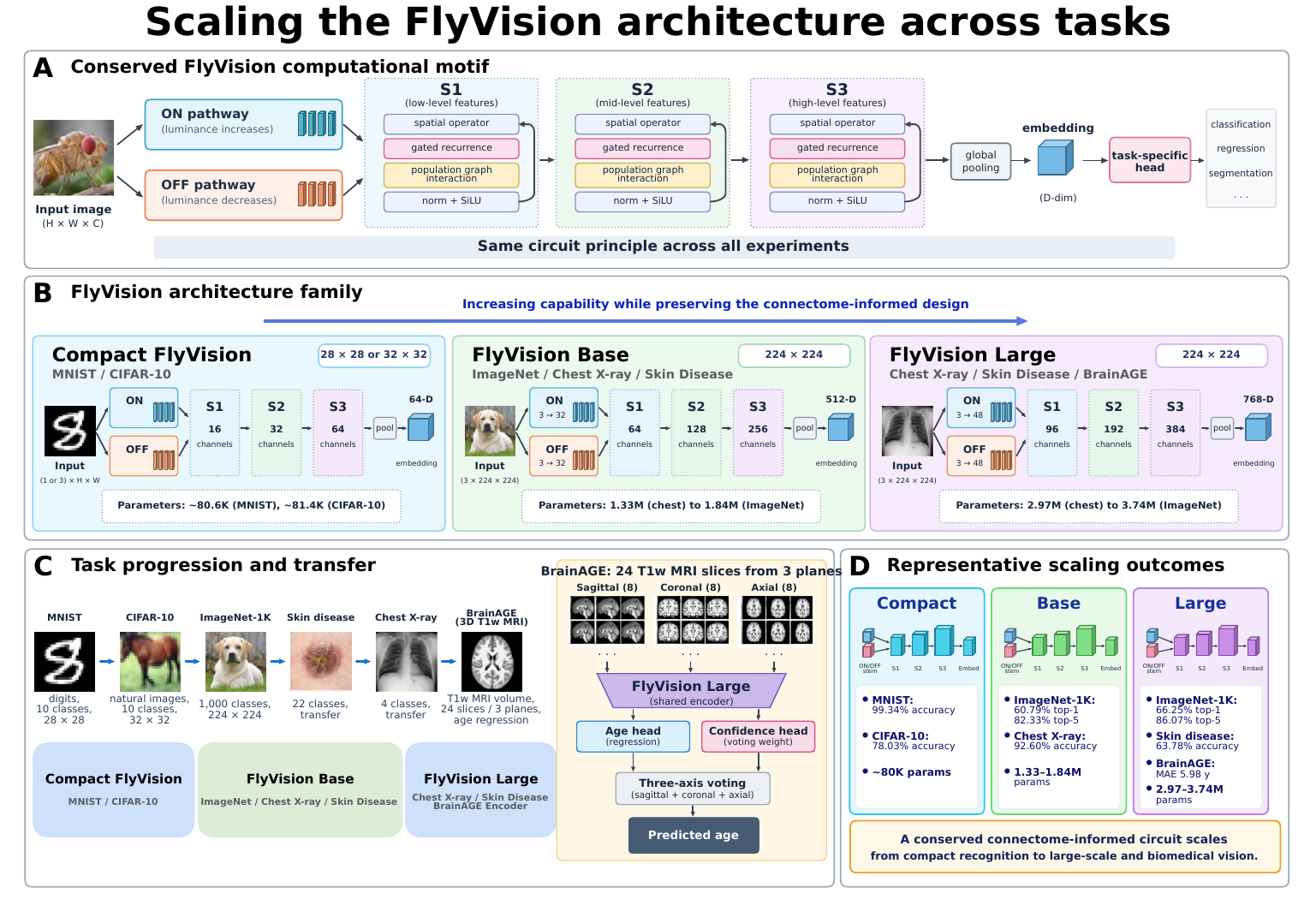}
\caption{Scaling the FlyVision architecture across tasks. (A) Conserved FlyVision computational motif. Parallel ON and OFF pathways feed three hierarchical stages (S1--S3), each combining a spatial operator, gated recurrence, population-graph interaction and normalization/nonlinearity before global pooling, embedding and a task-specific prediction head. (B) FlyVision architecture family. Compact FlyVision uses 16/32/64-channel stages and a 64-dimensional embedding for MNIST and CIFAR-10; FlyVision Base uses 64/128/256-channel stages and a 512-dimensional embedding for 224$\times$224 inputs; FlyVision Large uses 96/192/384-channel stages and a 768-dimensional embedding. Parameter counts vary with the task interface, from approximately 80,000 parameters in the compact model to 1.33--1.84 million in Base and 2.97--3.74 million in Large. (C) Task progression and transfer. The study advances from MNIST and CIFAR-10 through ImageNet-1K, skin-disease classification and chest radiography to BrainAGE regression on a three-dimensional T1-weighted MRI volume. A shared FlyVision Large encoder processes eight sagittal, eight coronal and eight axial T1w slices; the age head predicts slice age, the confidence head supplies learned voting weights, and three-axis aggregation produces the scan-level estimate. (D) Representative outcomes across the architecture family, including paired ImageNet-1K validation top-1/top-5 accuracy of 60.79\%/82.33\% for Base global surround and 66.25\%/86.07\% for Large global surround, together with compact recognition, biomedical transfer and multi-view brain-age regression.}\label{fig:architecture}
\end{figure}

ConnectomeX addresses this question through FlyVision, a visual-system-specific architecture that separates a reusable connectome-informed backbone from task-dependent input and output interfaces. Figure~\ref{fig:connectome} summarizes the translation from the \textit{Drosophila} visual-connectome prior to a scalable artificial backbone and the sequence of cross-domain tests. The conserved computational motif combines parallel ON/OFF pathways, three recurrent graph-interaction stages, global pooling and a task-specific head. Capacity is increased systematically from a compact 16/32/64-channel model for MNIST and CIFAR-10 to 64/128/256-channel FlyVision Base and 96/192/384-channel FlyVision Large models for 224$\times$224 inputs while preserving the same circuit organization (Fig.~\ref{fig:architecture}). The study evaluates this architecture family across MNIST \cite{mnistdb}, CIFAR-10 \cite{krizhevsky2009}, ImageNet-1K \cite{deng2009,russakovsky2015}, two-dimensional biomedical classification and multi-view BrainAGE regression. The BrainAGE arm extends the framework from two-dimensional images to scan-level analysis of volumetric structural MRI. Brain age is commonly estimated from T1-weighted MRI \cite{franke2010,colefranke2017,cole2017deep}. BrainAGE uses the pretrained two-dimensional FlyVision Large encoder to process sagittal, coronal and axial slices from each three-dimensional T1w volume and aggregates slice-level estimates across the three anatomical planes, providing a multi-view route from a two-dimensional pretrained encoder to scan-level prediction on volumetric structural MRI.

The experiments form a linked progression across increasing task and domain complexity. MNIST tests compact end-to-end trainability and parameter efficiency; CIFAR-10 introduces natural-image variability and controlled transfer from MNIST; ImageNet-1K tests 1,000-class recognition and supplies pretrained checkpoints for downstream transfer; skin-disease imaging and chest radiography test the same representation under distinct biomedical acquisition and label regimes; and BrainAGE evaluates reuse of the ImageNet-pretrained encoder for continuous prediction from three-dimensional T1w MRI through multi-view slice aggregation. The architecture-scaling map in Fig.~\ref{fig:architecture} links these tasks to the Compact, Base and Large FlyVision variants and makes the transfer path from large-scale natural-image pretraining to biomedical applications explicit. Transfer behavior is evaluated as an empirical property of the source-target-architecture combination, consistent with prior work showing that feature transferability varies with task distance and model/training regime \cite{yosinski2014,kornblith2019}.

ImageNet-1K \cite{deng2009,russakovsky2015} is central to this progression because it serves both as a scaling benchmark and as the source of pretrained weights used in the biomedical experiments. FlyVision Base global-surround and FlyVision Large global-surround reached 60.79\%/82.33\% and 66.25\%/86.07\% top-1/top-5 accuracy, respectively, while the later Large local-k7 model with a learned low-frequency branch reached 66.53\%/86.31\%. Skin-disease classification then provides an intermediate two-dimensional biomedical domain, extending a well-established line of image-based dermatologic classification \cite{esteva2017}: on an exact-content-audited 22-class split with 1,538 held-out images, FlyVision Large local-k7 with a learned low-frequency branch reached 63.78\% accuracy and 95.28\% macro-AUROC, compared with 66.19\% and 95.96\% for ResNet18. These experiments complement the chest-radiography and BrainAGE analyses by testing how a common architectural prior behaves as dataset scale, acquisition physics, class structure and prediction objective change.

The central objective is to characterize the generality of the connectome-informed computation through predictive performance, parameter efficiency, computational cost, transfer behavior, calibration and structured error analysis. Paired transfer experiments quantify source-target compatibility, while hierarchical classification and multi-view voting resolve how task structure redistributes errors. Together, the design provides a quantitative framework for testing how connectome-derived circuit motifs scale across computer-vision tasks (Figs.~\ref{fig:connectome} and \ref{fig:architecture}).

\section*{Results}
\subsection*{A connectome-informed backbone separates reusable circuit organization from task interfaces}
FlyVision implements a compact hierarchy in which two learned input branches process the image in parallel before information enters three recurrent graph-mixing stages. In the checkpoint-compatible two-dimensional implementation, each stage combines a feed-forward spatial operator, a gated recurrent operator and population-level interaction before normalization and nonlinear activation. The base model expands through 64, 128 and 256 channels to a 512-dimensional embedding; FlyVision Large uses 96, 192 and 384 channels with a 768-dimensional embedding. Global pooling and a task-specific head map the shared representation to the required output space. This design permits the recurrent and population-structured computation to remain stable while the number of input channels, spatial operator and prediction head are adapted to the dataset.

The architectural separation is important for the cross-domain experiments. MNIST and CIFAR-10 use compact 2D instantiations, ImageNet-1K trains the same design at 224$\times$224 resolution, and the radiography models load all 54 tensors from the ImageNet FlyVision checkpoints into checkpoint-compatible downstream reconstructions before replacing the 1,000-class classifier. BrainAGE likewise preserves the ImageNet-pretrained FlyVision Large encoder and extends it to scan-level volumetric neuroimaging through multi-view analysis. Each three-dimensional T1-weighted MRI volume is represented by eight sagittal, eight coronal and eight axial slices; each slice is embedded independently, and slice-level predictions are fused within and across anatomical views. Figure~\ref{fig:connectome} defines the connectome-derived design prior and cross-domain scope; Fig.~\ref{fig:architecture} then details the conserved computational motif, capacity scaling across Compact, Base and Large variants, transfer progression and representative outcomes.

\subsection*{Compact recognition on MNIST}
\begin{figure}[!htbp]\centering
\includegraphics[width=0.86\textwidth]{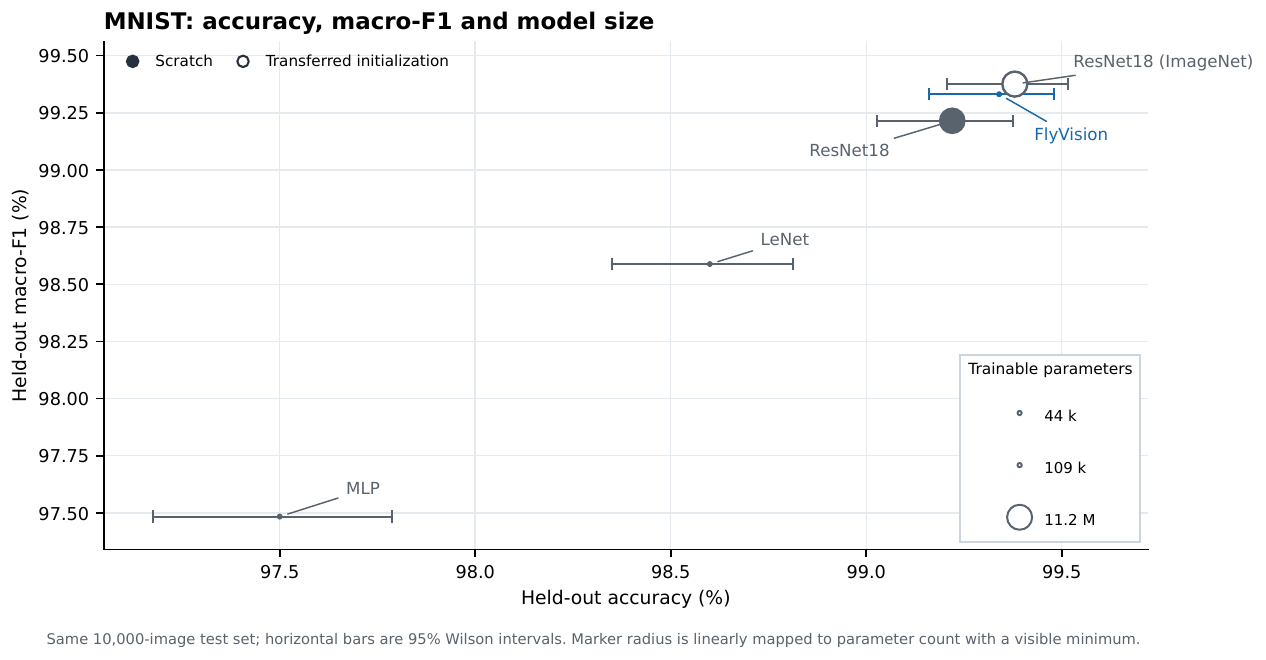}
\caption{MNIST held-out performance in the same overview format used for the larger classification tasks. Accuracy is plotted against macro-F1 for all benchmarked models on the common 10,000-image test set; horizontal bars show 95\% Wilson confidence intervals for accuracy. Filled markers indicate scratch training and open markers transferred initialization. Marker radius is linearly mapped to trainable parameter count, with a small minimum for visibility; exact counts are reported in Table~\ref{tab:mnist}.}\label{fig:mnist-overview}
\end{figure}

\begin{figure}[!htbp]\centering
\includegraphics[width=0.86\textwidth]{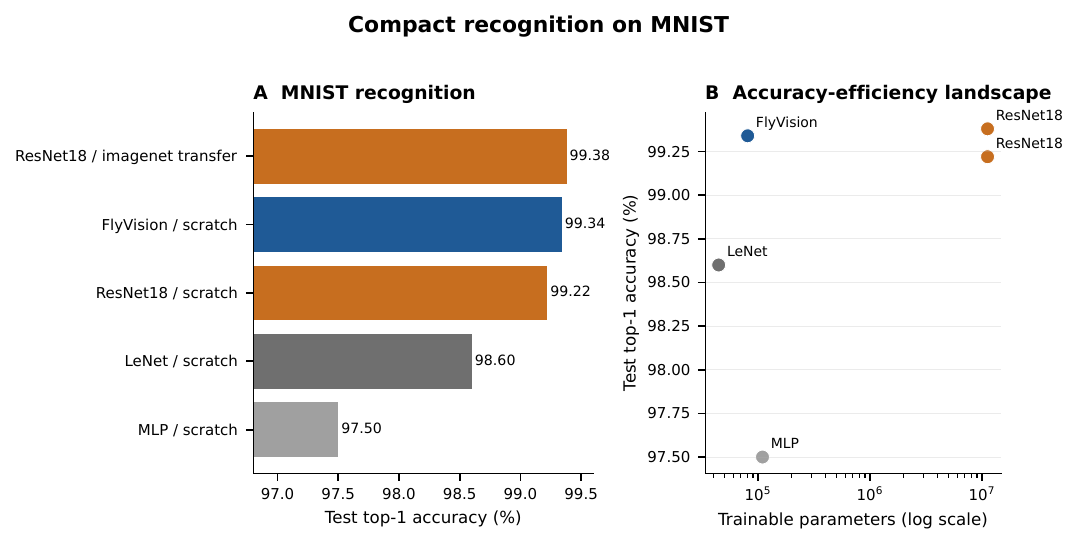}
\caption{Compact recognition on MNIST. (A) Held-out top-1 accuracy with 95\% Wilson confidence intervals for FlyVision and conventional baselines. (B) Accuracy plotted against trainable parameter count on a logarithmic axis. Scratch FlyVision reaches 99.34\% top-1 accuracy with 80,608 parameters, placing it near the highest observed accuracy while remaining markedly smaller than ResNet18.}\label{fig:mnist}
\end{figure}

MNIST \cite{mnistdb} provides the simplest controlled test of optimization and parameter efficiency. Under the frozen benchmark protocol, FlyVision trained from scratch reached 99.34\% top-1 accuracy and 99.33\% macro-F1 with 80,608 parameters. ImageNet-initialized ResNet18 reached 99.38\%, and scratch ResNet18 reached 99.22\%, both with 11,181,642 parameters. Thus, FlyVision operated within 0.04 percentage points of the highest observed accuracy, while ResNet18 contained approximately 139 times as many trainable parameters. LeNet reached 98.60\% and the MLP 97.50\% (Figs.~\ref{fig:mnist-overview} and \ref{fig:mnist}; Table~\ref{tab:mnist}).

The comparison separates predictive performance from model size. LeNet is smaller than FlyVision but has lower test accuracy, whereas ResNet18 is substantially larger with only a small absolute increase over FlyVision in the best initialization condition. The result therefore identifies a useful operating point for the connectome-informed architecture in a low-resolution regime: high accuracy, compact parameterization and end-to-end trainability.

\subsection*{Natural-image recognition reveals architecture-dependent transfer on CIFAR-10}
\begin{figure}[!htbp]\centering
\includegraphics[width=0.88\textwidth]{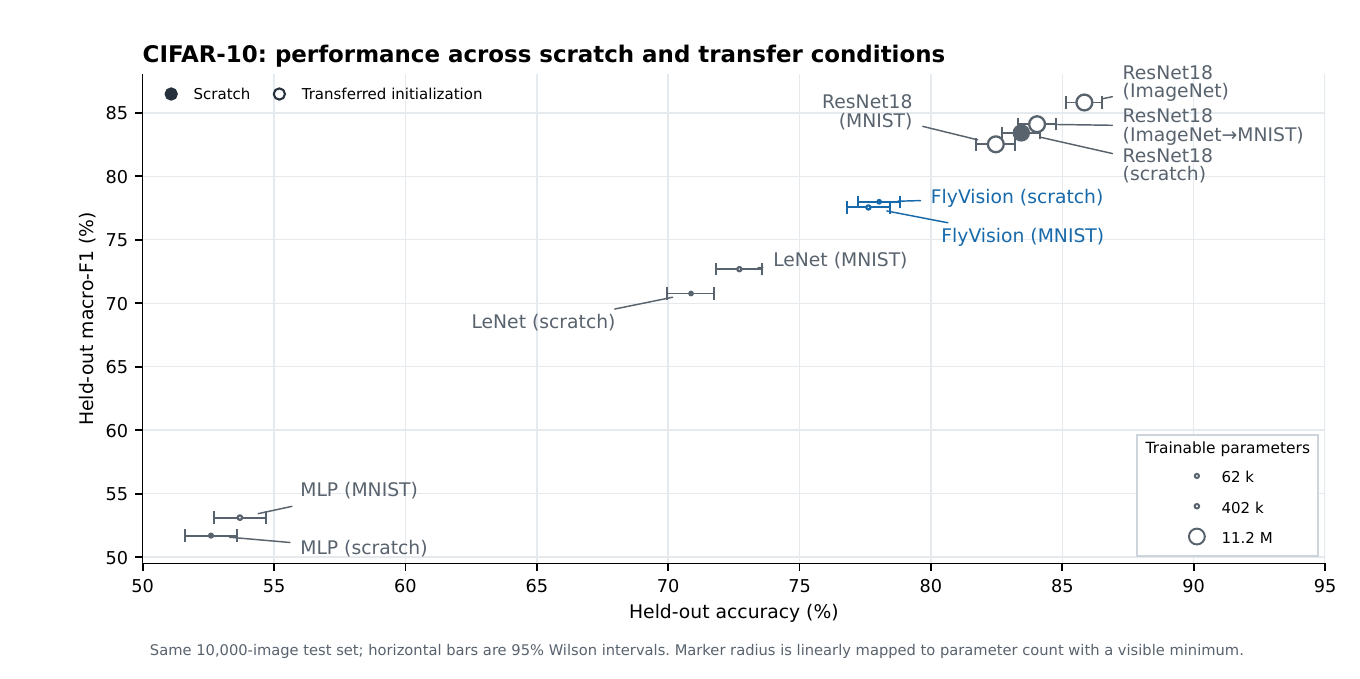}
\caption{CIFAR-10 performance across scratch and transfer conditions. Held-out accuracy is plotted against macro-F1 for all ten training conditions on the common 10,000-image test set; horizontal bars show 95\% Wilson confidence intervals for accuracy. Filled markers denote scratch training and open markers transferred initialization. Marker radius is linearly mapped to trainable parameter count, with a small minimum for visibility; exact counts and initialization routes are reported in Table~\ref{tab:cifar}.}\label{fig:cifar-overview}
\end{figure}

\begin{figure}[!htbp]\centering
\includegraphics[width=0.95\textwidth]{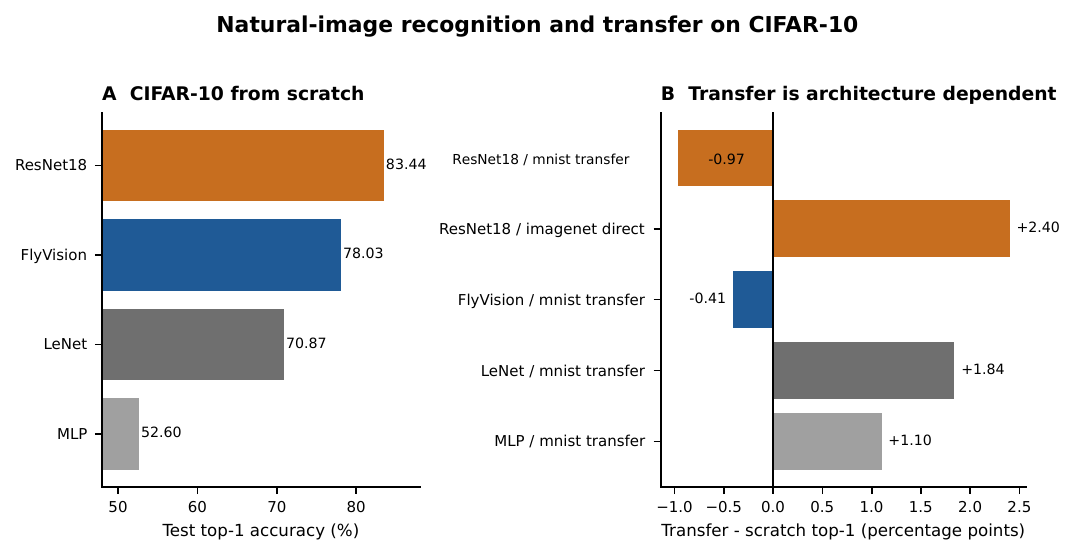}
\caption{Natural-image recognition and transfer on CIFAR-10. (A) Test top-1 accuracy for models trained from scratch. (B) Transfer-induced change in top-1 accuracy relative to each architecture's scratch baseline. Direct ImageNet initialization improves ResNet18 by 2.40 percentage points, whereas MNIST initialization changes FlyVision by -0.41 percentage points. The complete accuracy--macro-F1--parameter landscape across scratch and transferred conditions is shown in Fig.~\ref{fig:cifar-overview}.}\label{fig:cifar}
\end{figure}

CIFAR-10 \cite{krizhevsky2009} introduces color, object pose, texture and background variation while retaining a compact ten-class setting. From scratch, ResNet18 reached 83.44\% top-1 accuracy, FlyVision 78.03\%, LeNet 70.87\% and the MLP 52.60\% (Figs.~\ref{fig:cifar-overview} and \ref{fig:cifar}; Table~\ref{tab:cifar}). Direct ImageNet initialization increased ResNet18 to 85.84\%, providing the highest observed CIFAR-10 accuracy in the benchmark.

The paired transfer experiments show that source-task initialization is architecture dependent, consistent with prior evidence that feature transferability depends on source-target distance and the training regime \cite{yosinski2014,kornblith2019}. MNIST initialization changed FlyVision from 78.03\% to 77.62\% (-0.41 percentage points; 95\% CI, -1.22 to 0.40; Holm-adjusted exact McNemar $P=0.664$). The same source initialization changed ResNet18 by -0.97 points (adjusted $P=0.034$), LeNet by +1.84 points (adjusted $P=3.87\times10^{-4}$), and the MLP by +1.10 points (adjusted $P=0.085$). Direct ImageNet initialization increased ResNet18 from 83.44\% to 85.84\% (+2.40 points; 95\% CI, 1.69--3.11; adjusted $P=2.49\times10^{-10}$), whereas ImageNet-to-MNIST-to-CIFAR sequential initialization reached 84.04\%. Thus, transfer effects vary with source task, target task and architecture.

\subsection*{ImageNet-1K establishes large-scale visual recognition}
ImageNet-1K \cite{deng2009,russakovsky2015} tests whether the FlyVision design remains trainable as image diversity and label cardinality expand. At 224$\times$224 resolution, the selected validation checkpoints yielded 60.79\% top-1 and 82.33\% paired top-5 for FlyVision Base global surround (zero-based epoch 84), and 66.25\%/86.07\% for FlyVision Large global surround (epoch 82). Base local-k7 surround yielded 60.25\%/81.93\%. The corresponding 1,000-class global checkpoints contain 54 state-dict tensors and provide the initialization used in the downstream radiography experiments.

\begin{figure}[!htbp]\centering
\includegraphics[width=0.91\textwidth]{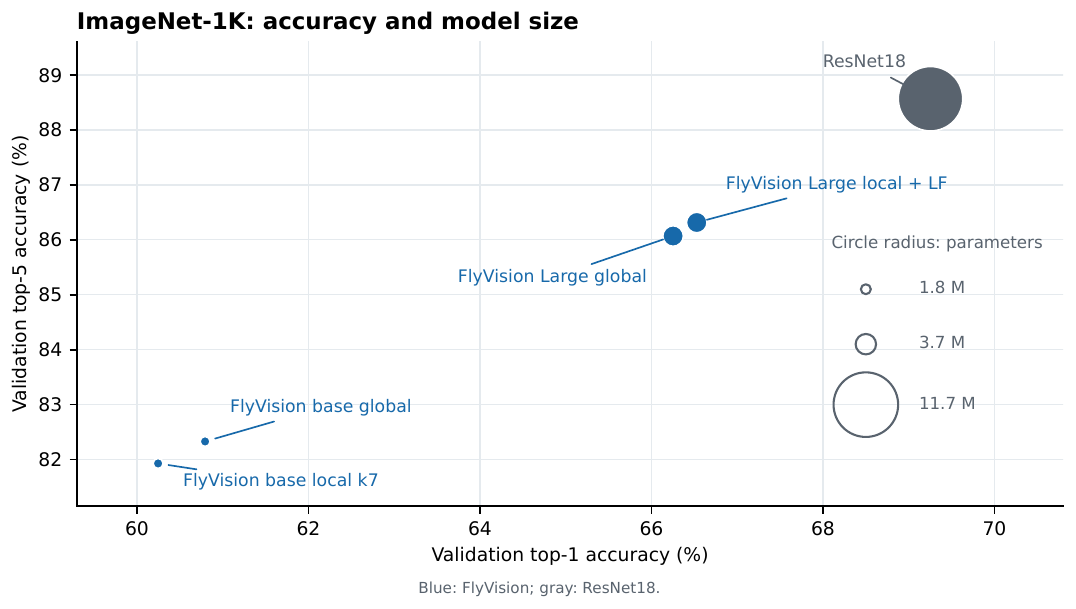}
\caption{ImageNet-1K validation performance and model size. Validation top-1 accuracy is plotted against paired top-5 accuracy for the completed models. Bubble radius is proportional to trainable parameter count. FlyVision variants are shown in blue and ResNet18 in gray. Here LF denotes the learned low-frequency pathway described in Methods. The Large local-k7/low-frequency checkpoint was evaluated on all 50,000 validation images. Exact top-1, top-5, parameter and available operation-count values are reported in Table~\ref{tab:imagenet}.}\label{fig:imagenet}
\end{figure}

Independent evaluation of a later FlyVision Large local-k7 model with a learned low-frequency branch produced 66.53\% top-1 and 86.31\% top-5 on the 50,000-image validation set; this 3.744-million-parameter checkpoint initialized the skin-disease FlyVision run. Scratch-trained ResNet18 reached 69.25\%/88.57\% under the shared ImageNet launcher. FlyVision Base global, Large global and ResNet18 contain 1.836, 3.737 and 11.690 million parameters, respectively; their measured forward Conv/Linear operation counts are 6.632, 14.803 and 3.628 GFLOPs for 224$\times$224 inputs (two FLOPs per multiply-accumulate). Thus, the ImageNet results separate parameter efficiency from computational cost: FlyVision uses substantially fewer parameters than ResNet18, while the recurrent and graph-interaction stages increase forward operation count in the profiled configurations (Fig.~\ref{fig:imagenet}; Table~\ref{tab:imagenet}).

\subsection*{Skin-disease imaging tests biomedical transfer outside radiography}
\begin{figure}[!htbp]\centering
\includegraphics[width=0.91\textwidth]{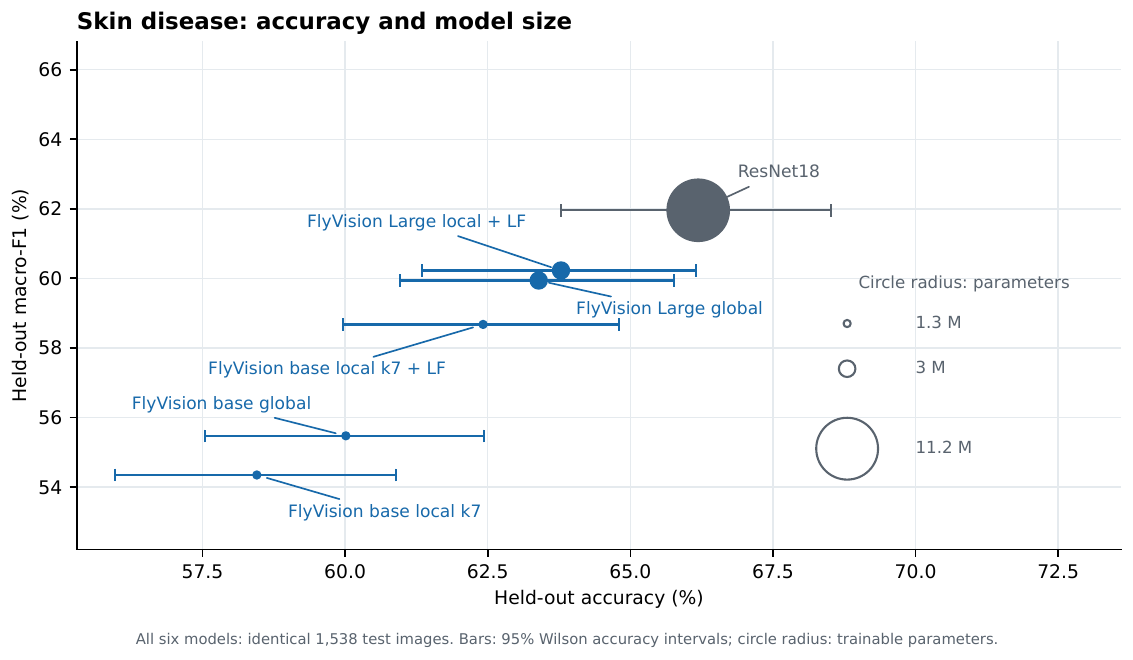}
\caption{Skin-disease held-out performance and model size. Test accuracy is plotted against macro-F1 for six models evaluated on the same 1,538 test images. Horizontal bars show 95\% Wilson intervals for accuracy, and bubble radius is proportional to trainable parameter count. Global and local-k7 refer to full-image and $7\times7$ neighborhood surrounds, respectively; LF is the learned low-frequency pathway. Full per-image probabilities support the one-versus-rest macro-AUROC values in Table~\ref{tab:skin} and the directional confusion analysis reported in the Results.}\label{fig:skin}
\end{figure}

Skin-disease classification provides a second two-dimensional biomedical setting with image statistics and label semantics distinct from natural photographs and chest radiographs. Image-based dermatologic classification has long provided a stringent test of visual representation quality because clinically relevant classes can differ by fine-grained morphology and color/texture cues \cite{esteva2017}. The SkinDisease image collection contains 15,444 source images in 22 labeled classes. Exact-content screening and contradictory-label quarantine yielded a prepared image-level split of 11,506 training, 2,042 validation and 1,538 test images. The six selected evaluation exports contain identical test sample keys, true labels and class mappings. Across five FlyVision variants, held-out accuracy ranged from 58.45\% (base local-k7) to 63.78\% (Large local-k7 with a learned low-frequency branch), and one-versus-rest macro-AUROC ranged from 94.43\% (base global) to 95.59\% (Large global). ResNet18 reached 66.19\% accuracy, 61.96\% macro-F1 and 95.96\% macro-AUROC. Table~\ref{tab:skin} reports all six models; Fig.~\ref{fig:skin} summarizes accuracy, macro-F1 and parameter count with 95\% Wilson accuracy intervals.

\begin{figure}[!htbp]\centering
\includegraphics[width=0.97\textwidth]{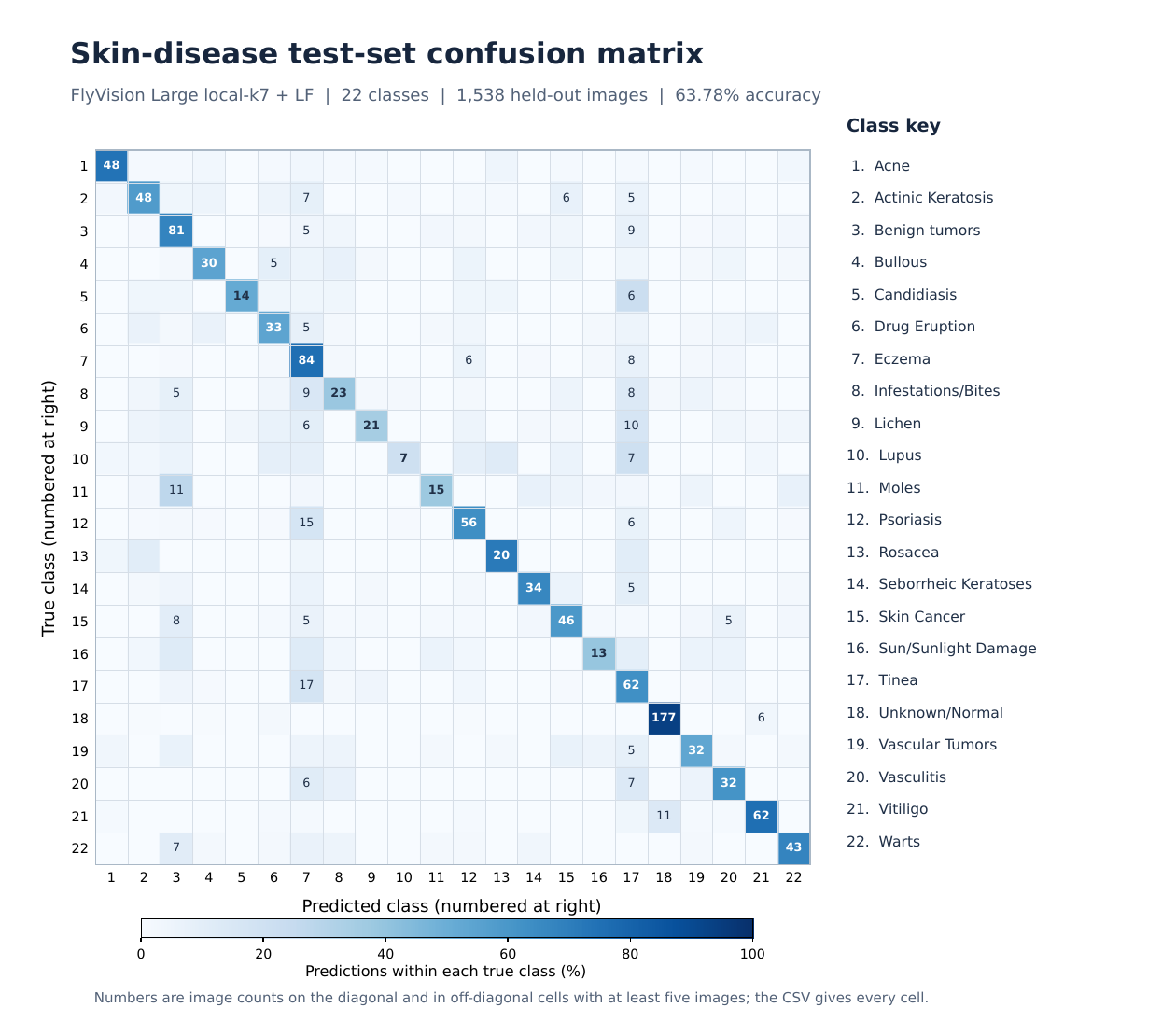}
\caption{Skin-disease held-out confusion matrix for FlyVision Large local-k7 with the learned low-frequency branch. The 22-class matrix uses the same 1,538 image-level test samples as Fig.~\ref{fig:skin}; rows are true classes and columns are predicted classes. Color indicates the percentage within each true-class row, while numerals give counts on the diagonal and for off-diagonal cells with at least five images. The accompanying CSV contains all cell counts and the paired per-image predictions. Overall accuracy is 981/1,538 (63.78\%). The evaluation therefore quantifies image-level generalization.}\label{fig:skin-confusion}
\end{figure}

Across the six models, the most recurrent directional confusions were eczema classified as psoriasis (6--16 of 111 eczema images), tinea classified as eczema (8--17 of 98), and benign tumors classified as skin cancer (4--17 of 120). Because every model was evaluated on the same held-out images, these counts describe paired model behavior on a common test set. Figure~\ref{fig:skin-confusion} shows the highest-accuracy FlyVision configuration's 22-class confusion matrix. The full matrices and classwise one-versus-rest AUROCs for all six models are included with the figure data. Five runs used resize/rotation jitter, whereas the Large local-k7/low-frequency run used spatial affine augmentation. Patient and lesion identifiers were not present in the available metadata; the prepared split is therefore image-level, and the reported estimates quantify held-out image-level generalization.

\subsection*{FlyVision occupies a favorable accuracy-parameter regime on four-class chest radiography}
The chest-radiography benchmark used the public COVID-19 Radiography Database \cite{chowdhury2020,rahman2021}, comprising 21,165 image-mask pairs: 3,616 COVID-19, 6,012 lung-opacity, 10,192 normal and 1,345 viral-pneumonia images. A deterministic class-stratified 70:15:15 image-level split produced 14,815 training, 3,175 validation and 3,175 held-out test images. The paired lung mask was applied in image space before ImageNet normalization.

\begin{figure}[!htbp]\centering
\includegraphics[width=0.88\textwidth]{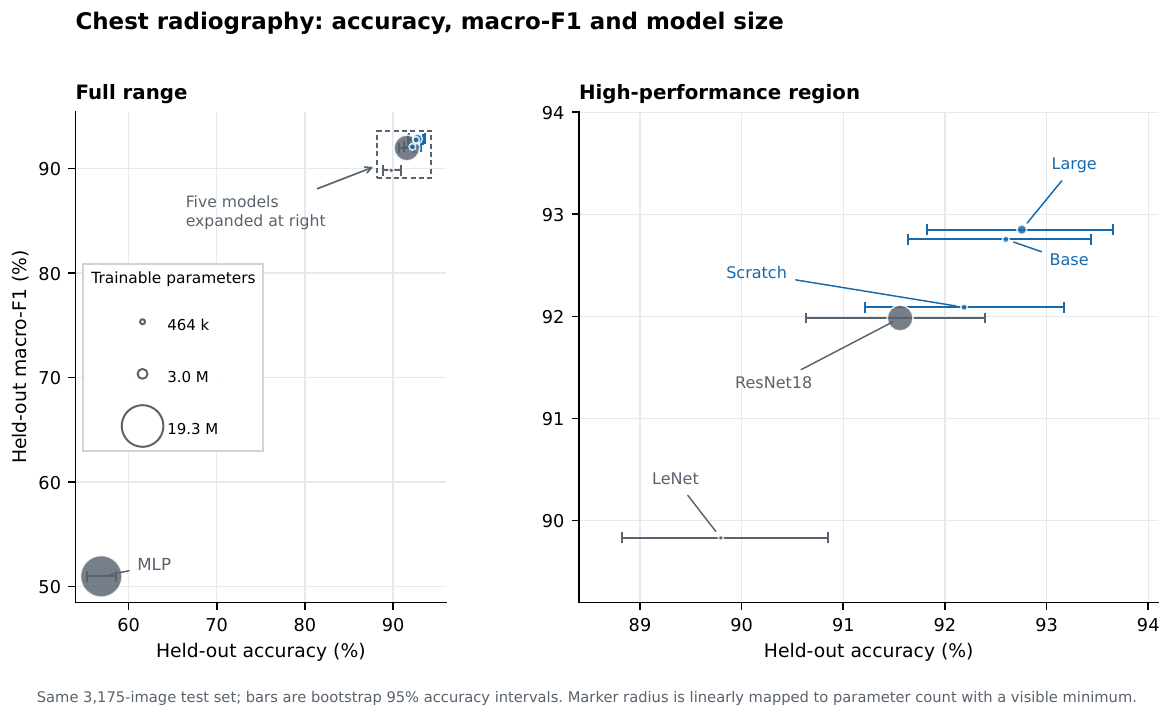}
\caption{Chest-radiography held-out performance in the cross-task overview format. Accuracy is plotted against macro-F1 for six models evaluated on the same 3,175-image test set; horizontal bars show bootstrap 95\% accuracy intervals. Marker radius is linearly mapped to trainable parameter count, with a small minimum for visibility. The five high-performing models cluster at the upper right of the full-range panel; the right panel labels and enlarges them using exact model coordinates and complete confidence intervals. The left panel retains the lower-performing MLP and the overall scale. Exact model size and performance values are reported in Table~\ref{tab:cxr}.}\label{fig:cxr-overview}
\end{figure}

\begin{figure}[!htbp]\centering
\includegraphics[width=0.82\textwidth]{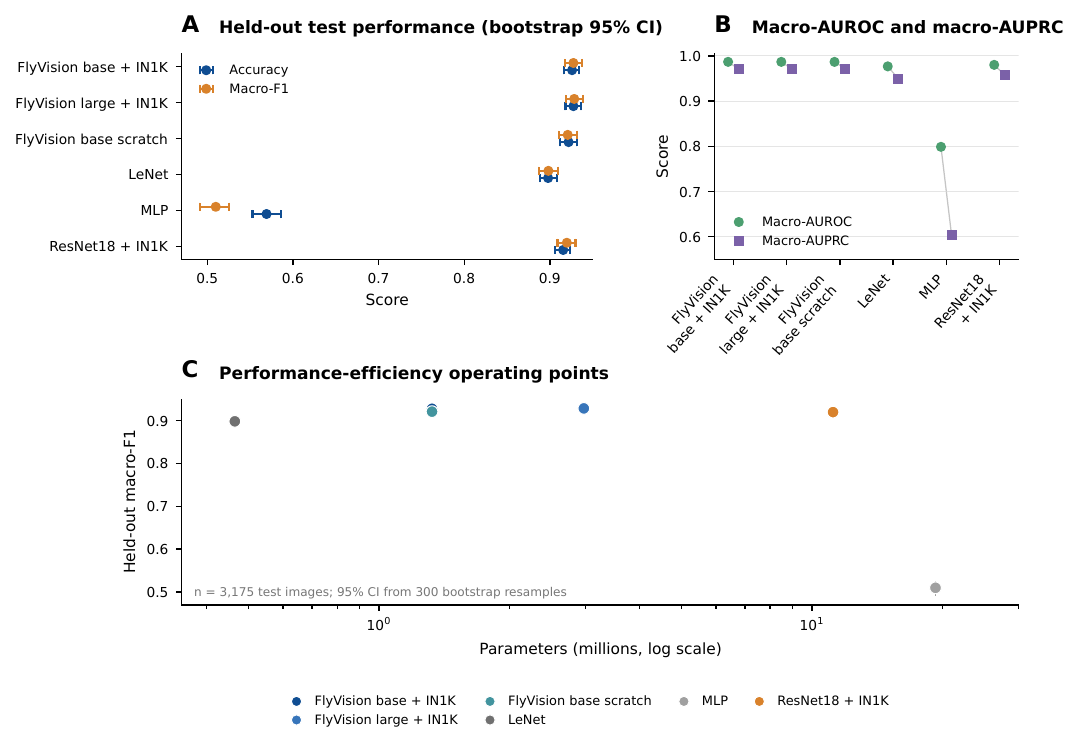}
\caption{Held-out chest-radiography benchmark. (A) Accuracy and macro-F1 for six models on the 3,175-image test set; horizontal bars show 95\% confidence intervals from 300 bootstrap resamples. (B) Test macro-AUROC and macro-AUPRC. (C) Macro-F1 versus total parameter count on a logarithmic x-axis; vertical bars show the same bootstrap macro-F1 intervals. IN1K, ImageNet-1K pretraining.}\label{fig:cxr-primary}
\end{figure}

On the held-out test set, FlyVision Large pretrained on ImageNet-1K achieved 92.76\% accuracy (95\% bootstrap CI, 91.83--93.66\%) and 92.85\% macro-F1 (91.84--93.91\%). FlyVision Base achieved 92.60\% accuracy (91.64--93.43\%) and 92.75\% macro-F1 (91.74--93.81\%), while the same base architecture trained from random initialization reached 92.19\% accuracy and 92.09\% macro-F1. ImageNet-pretrained ResNet18 reached 91.56\% accuracy and 91.98\% macro-F1; LeNet reached 89.80\% accuracy and 89.83\% macro-F1; the MLP reached 56.91\% accuracy and 50.98\% macro-F1 (Figs.~\ref{fig:cxr-overview} and \ref{fig:cxr-primary}; Table~\ref{tab:cxr}).

\begin{figure}[!htbp]\centering
\includegraphics[width=0.80\textwidth]{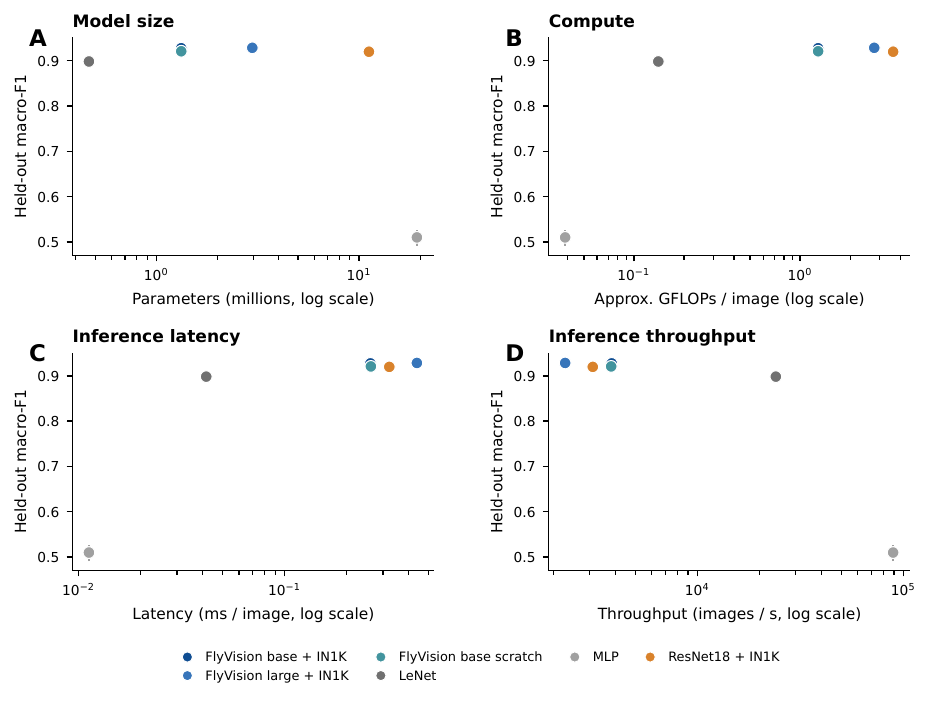}
\caption{Chest-radiography performance-efficiency trade-offs. Held-out macro-F1 is plotted against (A) total parameter count, (B) approximate GFLOPs per image measured from the executed chest-radiography forward graph, (C) single-GPU inference latency and (D) single-GPU throughput. The FlyVision radiography operation counts reflect its downstream stride/recurrent schedule; Table~\ref{tab:imagenet} reports the ImageNet-forward counts. X-axes are logarithmic; vertical lines show 95\% bootstrap confidence intervals for macro-F1.}\label{fig:cxr-efficiency}
\end{figure}

FlyVision Base contains 1.33 million parameters and requires approximately 1.28 GFLOPs per 224$\times$224 image in the executed chest-radiography forward graph, compared with 11.18 million parameters and approximately 3.63 GFLOPs for ResNet18. Within this downstream benchmark, ResNet18 therefore has 8.44 times as many parameters and requires 2.83 times as many approximate Conv/Linear FLOPs as FlyVision Base, while FlyVision Base point estimates are 1.04 percentage points higher for accuracy and 0.77 points higher for macro-F1. FlyVision Large contains 2.97 million parameters and requires approximately 2.79 GFLOPs in the same chest-radiography profiler. These downstream operation counts use the radiography execution schedule, whereas the ImageNet values in Table~\ref{tab:imagenet} use the ImageNet schedule. Within this benchmark, FlyVision Base occupies a favorable accuracy-parameter and downstream-compute operating region (Figs.~\ref{fig:cxr-overview}, \ref{fig:cxr-primary} and \ref{fig:cxr-efficiency}).

\begin{figure}[!htbp]\centering
\includegraphics[width=0.82\textwidth]{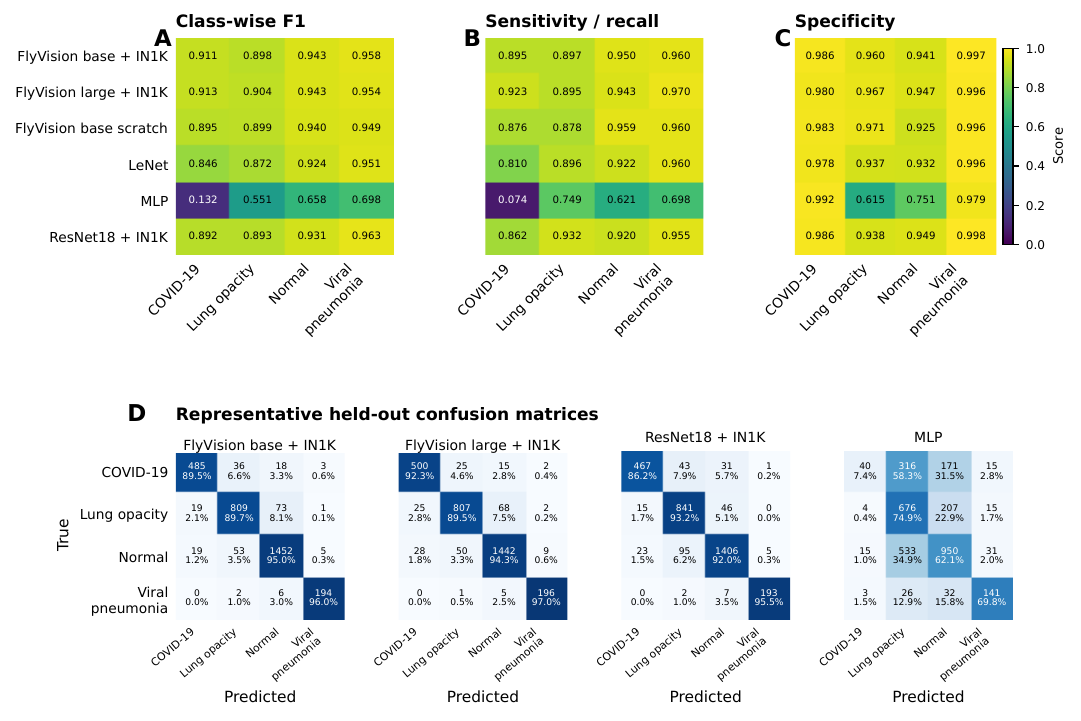}
\caption{Chest-radiography classwise behavior and error structure. (A--C) Test-set F1, sensitivity/recall and specificity for COVID-19, lung opacity, normal and viral pneumonia. (D) Representative confusion matrices for FlyVision Base, FlyVision Large, ResNet18 and MLP. Cells report observed counts and row-normalized percentages.}\label{fig:cxr-classwise}
\end{figure}

Classwise analysis further resolves where errors remain (Fig.~\ref{fig:cxr-classwise}). FlyVision Base classwise F1 spans 0.898 for lung opacity to 0.958 for viral pneumonia, whereas FlyVision Large spans 0.904 to 0.954. COVID-19 sensitivity is 0.895 for FlyVision Base and 0.923 for FlyVision Large, compared with 0.862 for ResNet18. Across the stronger models, the largest residual ambiguity occurs between lung opacity and normal radiographs, while viral pneumonia is comparatively well separated.

\subsection*{Hierarchical decomposition localizes the principal radiography errors}
\begin{figure}[!htbp]\centering
\includegraphics[width=0.96\textwidth]{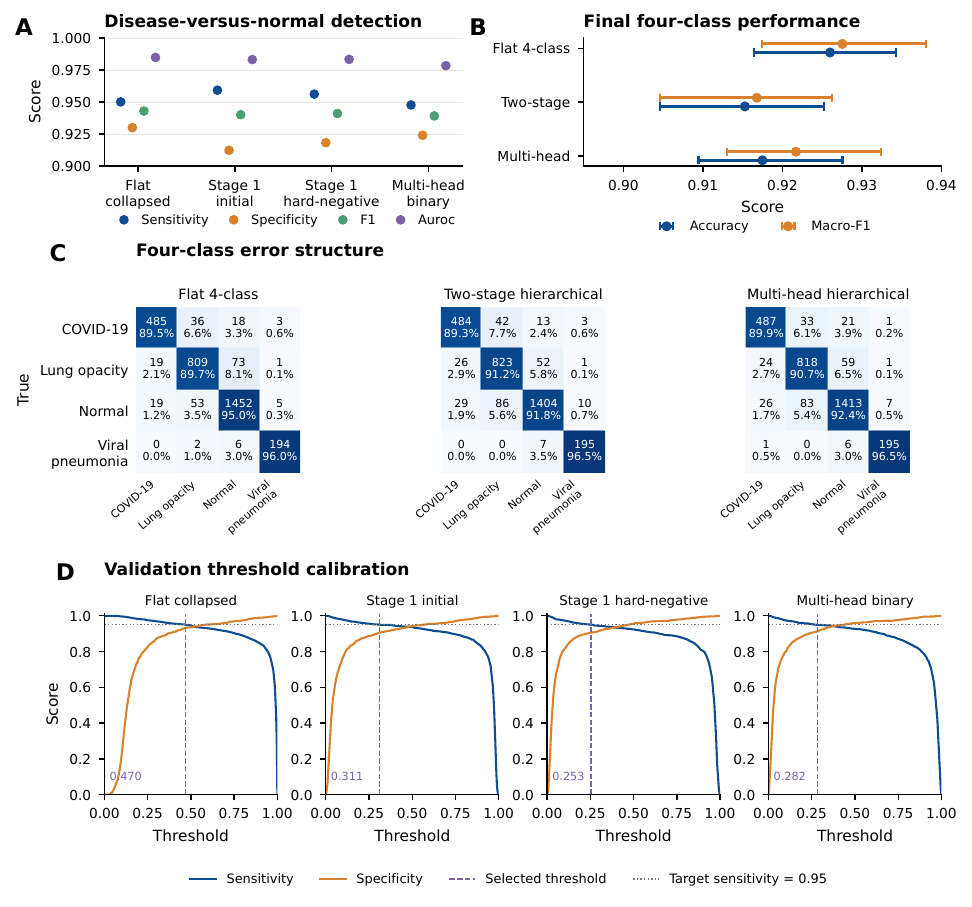}
\caption{Hierarchical chest-radiography analysis. (A) Disease-versus-normal sensitivity, specificity, F1 and AUROC for the collapsed flat classifier and hierarchical detectors. (B) Accuracy and macro-F1 with 95\% bootstrap confidence intervals for the final flat, two-stage and multi-head four-class systems. (C) Corresponding test confusion matrices. (D) Validation sensitivity-specificity curves with selected thresholds and the target disease sensitivity of 0.95.}\label{fig:cxr-hier}
\end{figure}

We reorganized radiography classification into two hierarchical formulations: a two-stage system that first separates normal from any disease and then classifies disease subtype, and a shared-backbone multi-head system with binary and subtype outputs. The flat FlyVision four-class model achieved 92.60\% accuracy and 92.75\% macro-F1. The two-stage hierarchy reached 91.53\% accuracy and 91.68\% macro-F1, and the multi-head hierarchy reached 91.75\% accuracy and 92.17\% macro-F1 (Fig.~\ref{fig:cxr-hier}; Table~\ref{tab:hier}).

\begin{figure}[!htbp]\centering
\includegraphics[width=0.90\textwidth]{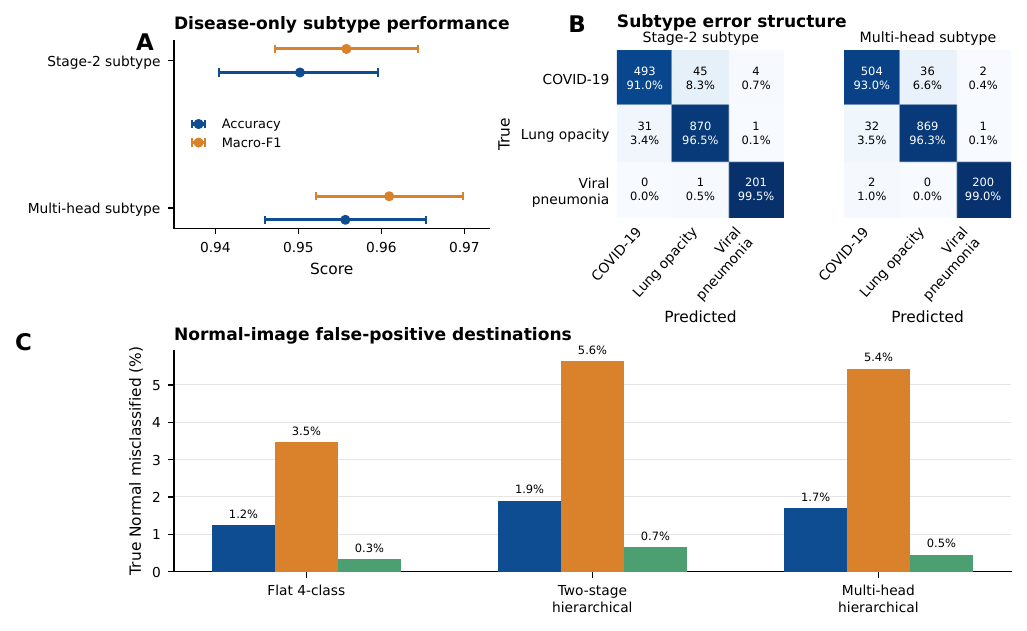}
\caption{Disease-subtype performance and radiography routing errors. (A) Accuracy and macro-F1 with 95\% bootstrap confidence intervals for the disease-only stage-2 classifier and the multi-head subtype output. (B) Subtype confusion matrices. (C) Percentage of true-normal test images routed to each disease class by the flat, two-stage and multi-head systems. Blue, orange and green bars denote COVID-19, lung opacity and viral pneumonia, respectively; lung opacity is the dominant false-positive destination for true Normal images in all three schemes.}\label{fig:cxr-subtype}
\end{figure}

The decomposition is most informative when the intermediate tasks are examined directly. Collapsing the flat model to normal-versus-disease yielded 95.02\% sensitivity, 93.00\% specificity, an AUROC of 0.9849 and an AUPRC of 0.9877. The dedicated disease-only subtype model achieved 95.02\% accuracy and 95.58\% macro-F1, while the multi-head subtype output achieved 95.57\% accuracy and 96.09\% macro-F1. End-to-end degradation is concentrated in the routing stage, while disease-only subtype discrimination remains strong. True normal images are most frequently routed toward lung opacity, identifying a specific error pathway that is obscured by aggregate four-class accuracy (Figs.~\ref{fig:cxr-hier} and \ref{fig:cxr-subtype}).

\subsection*{Multi-view FlyVision extends the pretrained visual representation to volumetric brain-age regression}
The BrainAGE experiment applies the pretrained FlyVision Large image encoder to scan-level age prediction from T1-weighted structural MRI. The 2,851-scan cohort is drawn from the same 13 source neuroimaging datasets described below, with participant-grouped 70:15:15 partitioning used to keep repeated scans from one participant within a single split; the held-out test split contains \BrainTestN{} scans. Each T1w volume is represented by eight sagittal, eight coronal and eight axial slices sampled across the brain extent, with selected slices center-padded to a common canvas to preserve in-plane geometry.

Each of the 24 slices is encoded by the ImageNet-pretrained FlyVision Large backbone after strict loading of all 54 checkpoint tensors and removal of the original 1,000-class classifier. A slice-level regression head produces an age estimate and a learned confidence score. Within each anatomical axis, the confidence-modulated vote is multiplied by a Gaussian positional prior centered in the brain extent ($\sigma=0.23$), normalized across the eight slices and used to obtain an axis-level age estimate. The final scan-level BrainAGE is the equal-weight mean of the sagittal, coronal and axial votes.

\BrainResultsParagraph
Descriptively, three-axis fusion reduced MAE by 1.45--1.50 years relative to the three individual-axis estimates.

\subsection*{Cross-task synthesis}
Across the study, the same circuit principle remains trainable from 28$\times$28 grayscale digits through 32$\times$32 natural images and 224$\times$224 large-scale or biomedical images, and the ImageNet-pretrained FlyVision Large representation is reused across anatomical slices for scan-level brain-age regression from three-dimensional T1w MRI. The benchmark sequence resolves distinct properties of the architecture. MNIST establishes compact high-accuracy recognition; CIFAR-10 reveals that transfer depends on source-target compatibility; ImageNet-1K demonstrates scaling to 1,000 classes and supplies reusable pretrained checkpoints; skin-disease classification and chest radiography test two different biomedical image regimes; hierarchical radiography localizes routing errors; and BrainAGE extends the same pretrained representation to continuous prediction through multi-view aggregation. Across these settings, parameter count, computation, predictive quality and error structure provide complementary views of how the connectome-informed design scales.

\section*{Discussion}
ConnectomeX evaluates whether organizational principles derived from the Drosophila visual connectome can function as reusable architectural constraints for artificial vision. The central result is the persistence of one computational motif across progressively different regimes: compact digit recognition, natural-image classification, 1,000-class ImageNet recognition, two biomedical classification problems and a multi-view volumetric structural-MRI extension for BrainAGE. The architecture scales capacity while preserving parallel input pathways, recurrent processing and population-level interaction, allowing task performance to be interpreted in relation to a shared circuit design.

The low-resolution experiments distinguish compactness from transferability. FlyVision reaches 99.34\% accuracy on MNIST with 80,608 parameters, placing it near the highest observed accuracy while using roughly two orders of magnitude fewer parameters than ResNet18. On CIFAR-10, scratch FlyVision reaches 78.03\%, below ResNet18 but above LeNet and the MLP. MNIST initialization changes FlyVision accuracy by -0.41 percentage points, whereas the same source task improves LeNet and the MLP and direct ImageNet initialization improves ResNet18. This architecture-dependent pattern is consistent with the broader transfer-learning literature: feature reuse depends on both source-target relatedness and the representation learned by a particular architecture/training regime \cite{yosinski2014,kornblith2019}.

ImageNet-1K establishes the large-scale operating regime of the architecture family. Increasing FlyVision capacity from Base to Large raises top-1 accuracy from 60.79\% to 66.25\% for the global-surround models, and the Large local-k7/low-frequency model reaches 66.53\%. ResNet18 remains higher at 69.25\%, but with 11.690 million parameters versus approximately 3.74 million for FlyVision Large. At the same time, the profiled FlyVision models require more Conv/Linear operations than ResNet18, emphasizing that parameter efficiency and computational efficiency are distinct design properties. This distinction is important for connectome-informed networks because recurrence and population interaction can increase computation without proportionally increasing parameter count.

The two-dimensional biomedical experiments test transfer under different acquisition physics and label structures. The skin experiment extends image-based dermatologic classification into a 22-class setting \cite{esteva2017}, whereas the radiography benchmark uses a thoracic imaging distribution and a distinct four-class label space \cite{chowdhury2020,rahman2021}. In the 22-class skin-disease benchmark, FlyVision Large local-k7 with a learned low-frequency branch reaches 63.78\% accuracy, 60.23\% macro-F1 and 95.28\% macro-AUROC with 2.992 million parameters, compared with 66.19\%, 61.96\% and 95.96\% for the 11.188-million-parameter ResNet18. In chest radiography, ImageNet-pretrained FlyVision Base and Large reach 92.60\% and 92.76\% accuracy with 1.33 and 2.97 million parameters, respectively, and both occupy a favorable accuracy-parameter regime relative to ResNet18. The radiography hierarchy further shows that disease-only subtype discrimination is stronger than the final four-class pipeline, localizing a substantial fraction of residual error to the normal-versus-disease routing decision.

The BrainAGE extension applies the same representation to a different prediction objective and to scan-level analysis of volumetric structural MRI. Brain age is commonly estimated by learning a mapping from T1-weighted neuroimaging to chronological age \cite{franke2010,colefranke2017,cole2017deep}. The BrainAGE model applies the ImageNet-pretrained FlyVision Large encoder independently to eight sagittal, eight coronal and eight axial slices, with the resulting age estimates combined by confidence-modulated Gaussian voting within each plane and equal-weight fusion across planes. On the held-out T1w test set, the fused prediction reached an MAE of \BrainMAE{} years and $R^2=\BrainRtwo$, compared with axis-specific MAEs of \BrainSagMAE{}, \BrainCorMAE{} and \BrainAxMAE{} years. This design uses the pretrained two-dimensional representation for scan-level regression through multi-view aggregation of a three-dimensional T1w volume.

The cross-domain results motivate direct component-level tests of the connectome-derived design. Sparse topology, recurrent pathways, ON/OFF segregation, population interaction and connection sign can be perturbed independently while matching parameter count and optimization budget. The benchmark suite provides quantitative reference conditions for such perturbations through aggregate performance, paired transfer effects, classwise errors, hierarchical routing, calibration and multi-view fusion.

The data partitions define the level at which biomedical generalization is measured. The skin collection supports exact-content duplicate screening but lacks patient or lesion identifiers, and the public radiography release is evaluated with deterministic image-level stratification; both therefore report held-out image-level performance. BrainAGE uses participant-grouped partitioning so repeated scans from a participant remain within one split. Patient-, site- and scanner-separated evaluation would provide a complementary test of deployment-level generalization.

Taken together, the results establish a coherent interface between connectomics and machine learning: biological circuit structure supplies experimentally grounded architectural hypotheses, and cross-domain artificial-vision benchmarks provide a controlled setting in which those hypotheses can be scaled, transferred and perturbed. The resulting program treats the connectome as a source of testable computational structure, linking circuit organization to measurable properties of modern vision systems.

\section*{Methods}
\subsection*{Study design and reproducibility}
The study combines deterministic MNIST and CIFAR-10 benchmarks, ImageNet-1K pretraining, two biomedical classification settings, hierarchical chest-radiography evaluation and a volumetric T1-weighted-MRI BrainAGE extension. Analyses are generated from fixed experiment outputs, including checkpoint metadata, aggregate metric tables and, where permitted by the applicable data-use terms, per-sample prediction files. Internal manuscript-reproduction resources retained by the authors contain processed tables used for the figures, available split and prediction artifacts, figure-recreation code and model-performance summaries.

\subsection*{FlyVision architecture}
The checkpoint-compatible two-dimensional FlyVision implementation contains a dual ON/OFF stem, three recurrent graph-mixing stages, adaptive global pooling, an embedding layer and a task-specific prediction head. In the ON/OFF stem, the surround is the per-channel spatial mean over the full image (global) or a reflect-padded $7\times7$ local mean at each pixel (local-k7). The ON and OFF responses are $\mathrm{ReLU}(x-s)$ and $\mathrm{ReLU}(s-x)$, respectively, for input $x$ and surround $s$; ``+ LF'' denotes a zero-initialized, learnable $5\times5$ stride-2 convolution of the local surround map added to the stem output as a low-frequency pathway. The ImageNet Base/Large implementation represented in Fig.~\ref{fig:connectome} uses stage strides 1/2/2 after the stride-2 stem and one/two/two recurrent updates across S1/S2/S3. In the chest-radiography implementation, the same 54-tensor checkpoint schema is loaded into a downstream reconstruction with stage strides 2/2/2 and one recurrent update per stage. Because stride and recurrent-update count are execution hyperparameters rather than state-dictionary tensors, the checkpoint-compatible ImageNet and radiography models execute different forward schedules, accounting for the distinct GFLOP counts reported in Tables~\ref{tab:imagenet} and \ref{tab:cxr}. The chest-radiography base model uses two independent 3-to-32-channel 5$\times$5 convolutions in the stem and stage widths of 64, 128 and 256 channels followed by a 512-dimensional embedding. FlyVision Large uses 3-to-48-channel stem branches, stage widths of 96, 192 and 384 channels and a 768-dimensional embedding. Each stage combines a feed-forward 3$\times$3 convolution, a 3$\times$3 recurrent convolution and a four-population graph mixer; learnable scalar gates modulate the recurrent and graph contributions, GroupNorm provides channel normalization \cite{wu2018}, and SiLU/Swish provides the activation nonlinearity \cite{ramachandran2018}.

The MNIST/CIFAR-10 implementation uses the same organizing motifs at smaller channel widths. Its FlyVision network contains an ON/OFF stem followed by three recurrent stages with widths 16, 32 and 64, recurrent-step configuration 1/2/2, graph interactions and lateral competition, adaptive pooling and a 64-dimensional embedding. The MNIST model contains 80,608 trainable parameters and the CIFAR-10 model 81,408, reflecting the different input channels and classifier interfaces.

\subsection*{MNIST and CIFAR-10 datasets}
MNIST \cite{mnistdb} contains 60,000 training and 10,000 test grayscale digit images. The benchmark reserves 6,000 images from the official training set for validation using a deterministic class-stratified split. CIFAR-10 \cite{krizhevsky2009} contains 50,000 training and 10,000 test RGB images across ten object classes; 5,000 training images are reserved for validation using the same deterministic stratified procedure. Dataset normalization statistics are estimated from the training subset only. CIFAR-10 training uses random 32$\times$32 crops with four-pixel padding and random horizontal flips; validation and test transforms are deterministic.

\subsection*{MNIST/CIFAR-10 optimization, transfer and evaluation}
All benchmark runs use 100 epochs, batch size 128, AdamW \cite{loshchilov2019} with learning rate $10^{-3}$ and weight decay $10^{-4}$, cosine decay to $10^{-5}$, and CUDA automatic mixed precision when available. FlyVision uses a gradient-clipping norm of 5 after AMP unscaling. Validation and test evaluation are performed in FP32. The checkpoint system stores model, optimizer, scheduler, GradScaler, random-number-generator state, DataLoader generator, history and protocol signature. Final reporting includes top-k accuracy, macro and weighted classification metrics, one-versus-rest AUROC, Brier score, expected calibration error, Wilson confidence intervals, inference timing and parameter counts.

For CIFAR-10 transfer runs, compatible learned states are initialized from completed MNIST source runs and the target classifier is reset. ResNet18 is additionally evaluated with direct torchvision ImageNet-1K initialization and with ImageNet-to-MNIST-to-CIFAR sequential initialization. Scratch-versus-transfer comparisons use paired per-image predictions and exact McNemar tests with Holm adjustment across planned comparisons.

\subsection*{ImageNet-1K pretraining}
The global-surround FlyVision ImageNet-1K checkpoints were trained from scratch on ImageNet-1K \cite{deng2009,russakovsky2015} at 224$\times$224 resolution for 90 scheduled epochs with seed 1102. Their best-top-1 metadata are recorded at zero-based epochs 84 (base) and 82 (Large); ResNet18 is at epoch 89. The common single-GPU launcher uses batch size 128, AdamW (initial learning rate $10^{-3}$, weight decay $10^{-4}$), five linear warm-up epochs and cosine decay to $10^{-6}$, label smoothing 0.1, random resized crop, horizontal flip, RandAugment \cite{cubuk2020}, random erasing probability 0.1, ImageNet normalization, BF16 training and FP32 validation. FlyVision uses gradient clipping at norm 5; ResNet18 does not. The later Large local-k7/low-frequency run manifest records one NVIDIA RTX PRO 6000 Blackwell GPU and the same recipe. Best-top-1 checkpoints are reported with their paired top-5 values; the later local-k7/low-frequency checkpoint was independently evaluated on all 50,000 validation images. 

\subsection*{Skin-disease dataset and evaluation}
The skin-disease benchmark used the Skin Disease Dataset published on Kaggle by Prashant Kumar Mishra \cite{mishraSkinDisease}. The downloaded collection contained 15,444 images assigned to 22 source classes: Acne, Actinic Keratosis, Benign tumors, Bullous, Candidiasis, Drug Eruption, Eczema, Infestations/Bites, Lichen, Lupus, Moles, Psoriasis, Rosacea, Seborrheic Keratoses, Skin Cancer, Sun/Sunlight Damage, Tinea, Unknown/Normal, Vascular Tumors, Vasculitis, Vitiligo and Warts. The files were distributed in train/test directories organized by class; directory names served as the supplied labels, and the final split assignments are recorded in the packaged split manifest. The downloaded release did not provide patient or lesion identifiers or documentation of how the original labels were assigned; accordingly, the present benchmark quantifies image-level generalization. SHA-256 screening found 306 duplicate-content groups, 37 groups with contradictory labels and 63 groups crossing the original train/test boundary. The preparation code quarantined conflicting-label hashes, removed training copies of held-out test content, retained one representative per same-label hash, and created a seed-42 class-stratified validation subset from the remaining training pool. The prepared image-level split was 11,506/2,042/1,538 for train/validation/test; patient or lesion grouping was unavailable. All six evaluations used the prepared 1,538-image test split, 224$\times$224 RGB images, ImageNet mean/standard-deviation normalization and a common 22-class index. The five FlyVision runs and ResNet18 used full-model AdamW fine-tuning with learning rate $10^{-4}$, weight decay $10^{-4}$, batch size 64, seed 42, at most 100 epochs and validation-top-1 checkpoint selection. The base-global, base-local-k7, base-local-k7/low-frequency, Large-global and ResNet18 manifests record resize/rotation jitter; Large local-k7/low-frequency records spatial affine augmentation. FlyVision runs used gradient clipping at norm 5; ResNet18 did not. Initialization provenance was verified from the run manifests for analyses that specifically compare ImageNet transfer conditions; all six models contribute to held-out predictive benchmark summaries. Accuracy intervals are 95\% Wilson intervals. For each class, AUROC was calculated one-versus-rest from the saved softmax probability column and binary true label; macro-AUROC is the unweighted mean of the 22 class AUROCs. Confusion counts use argmax predictions. The six prediction files were checked against the packaged test manifest and each other for identical sample keys and true labels. The six held-out prediction files are evaluated on the same fixed test set, enabling direct descriptive comparison of per-image errors and ranking metrics.

\subsection*{Chest-radiography dataset and split}
The radiography benchmark uses the COVID-19 Radiography Database \cite{chowdhury2020,rahman2021}. The indexed release contains 21,165 image-mask pairs: 3,616 COVID-19, 6,012 lung-opacity, 10,192 normal and 1,345 viral-pneumonia images. Images are read as three-channel tensors and masks as single-channel tensors. A deterministic class-stratified split with random seed 2026 assigns 70\% of images to training (n=14,815), 15\% to validation (n=3,175) and 15\% to testing (n=3,175).

\subsection*{Chest-radiography preprocessing and optimization}
Images and paired lung masks are resized to 224$\times$224 pixels. During training, paired spatial augmentation uses left-right flipping, mild affine transformation and anisotropic resampling; intensity augmentation uses gamma variation, Gaussian noise and blur. Spatial transforms are applied jointly to image and mask with nearest-neighbor interpolation for the mask. The binary lung mask is applied in image space before ImageNet normalization. Validation and test images use deterministic resizing and masking without stochastic augmentation.

Flat four-class models are optimized with AdamW \cite{loshchilov2019} using an initial learning rate of 0.0003, weight decay 0.0001, five warm-up epochs and cosine decay to 0.000002. The loss is class-weighted cross-entropy with label smoothing of 0.05. Training uses mixed precision, a per-GPU batch size of 16 and a maximum of 200 epochs. Early stopping monitors validation macro-F1 with patience of ten epochs. Test metrics are computed from the saved best-validation checkpoint.

\subsection*{Radiography model initialization and baselines}
FlyVision Base and FlyVision Large load all 54 tensors from the corresponding ImageNet-1K checkpoints before the 1,000-class classifier is replaced with a four-class head. The downstream reconstruction preserves checkpoint tensor shapes and parameter values and uses stage strides 2/2/2 with one recurrent update per stage; the ImageNet graph uses stage strides 1/2/2 with recurrent-update counts 1/2/2. The scratch FlyVision control uses this same radiography downstream graph with random initialization. Conventional comparators are LeNet \cite{lecun1998}, an MLP and ImageNet-pretrained ResNet18 \cite{he2016}. Parameter counts are measured directly from trainable tensors. Multiply-accumulate operations are measured from the executed radiography forward graph for a 224$\times$224 input and reported as approximate GFLOPs using two floating-point operations per MAC. The reported radiography operation counts correspond to this downstream execution schedule; Table~\ref{tab:imagenet} reports the ImageNet schedule.

\subsection*{Hierarchical radiography models}
The two-stage hierarchy trains one FlyVision model for normal-versus-disease detection and a second model on disease-only images for three-way subtype classification among COVID-19, lung opacity and viral pneumonia. A hard-negative detector refinement oversamples difficult normal examples. The multi-head formulation shares a FlyVision backbone between binary and subtype heads. Detection thresholds are calibrated on validation predictions with a target disease sensitivity of 0.95. Final four-class predictions are reconstructed from the binary and subtype outputs and evaluated on the same held-out test split as the flat model.

\subsection*{BrainAGE cohort, preprocessing and multi-view voting}
The BrainAGE experiment performs scan-level age prediction from T1-weighted structural MRI. The 2,851-scan cohort is assembled from 13 source neuroimaging datasets: the Alzheimer's Disease Neuroimaging Initiative (ADNI); the Brain Genomics Superstruct Project (BGSP) \cite{holmes2015bgsp}; Neuroimaging in Frontotemporal Dementia (NIFD/FTLDNI); the Southwest University Longitudinal Imaging Multimodal (SLIM) study \cite{liu2017slim}; the Dallas Lifespan Brain Study (DLBS) \cite{park2025dlbs}; the Southwest University Adult Lifespan Dataset (SALD) \cite{wei2018sald}; the Australian Imaging, Biomarkers and Lifestyle Study of Ageing (AIBL) \cite{ellis2009aibl}; the Consortium for Reliability and Reproducibility (CoRR) \cite{zuo2014corr}; SchizConnect \cite{wang2016schizconnect}; IXI \cite{ixiDataset}; OASIS-1 \cite{marcus2007oasis}; OASIS-2 \cite{marcus2010oasis}; and the Parkinson's Progression Markers Initiative (PPMI) \cite{ppmi2011}. Data used in preparation of this article were obtained in part from the ADNI database (\href{https://adni.loni.usc.edu}{adni.loni.usc.edu}). ADNI was launched in 2003 as a public--private partnership led by Principal Investigator Michael W. Weiner, MD. Its original goal was to determine whether serial MRI, positron-emission tomography, other biological markers, and clinical and neuropsychological assessment could be combined to measure progression of mild cognitive impairment and early Alzheimer's disease; current goals include biomarker validation for clinical trials, improving cohort diversity and generalizability, and providing data on Alzheimer's disease diagnosis and progression to the scientific community. For up-to-date information, see \href{https://adni.loni.usc.edu}{adni.loni.usc.edu}. Participant-grouped 70:15:15 splitting with random seed 2026 keeps repeated scans from one participant within a single split; the held-out test set contains 433 scans. Sex is retained only for descriptive subgroup evaluation.

Preprocessing identifies the brain extent of each T1-weighted MRI volume and samples eight slices from each of the sagittal, coronal and axial planes between approximately 5\% and 95\% of that extent, yielding 24 slices per scan. Selected slices are center zero-padded to a common canvas to preserve native in-plane geometry, repeated across three channels and normalized for the ImageNet-pretrained encoder.

The model uses the FlyVision Large ImageNet checkpoint as a shared 2D encoder. All 54 checkpoint tensors are strictly loaded before the original classifier is replaced by an identity mapping, exposing the 768-dimensional embedding for each slice. A slice-age head consists of LayerNorm \cite{ba2016}, a 256-unit hidden layer with GELU activation \cite{hendrycks2016} and dropout of 0.20, and a scalar age output. A separate confidence head uses LayerNorm, a 128-unit hidden layer with tanh activation and produces a scalar confidence score $c_i$ for each slice. The complete BrainAGE model, including the shared FlyVision Large encoder and the age/confidence heads, contains 3.27 million trainable parameters. For normalized slice position $p_i$, the positional prior is $g(p_i)=\exp[-0.5((p_i-0.5)/0.23)^2]$, and the unnormalized slice weight is $g(p_i)(0.05+c_i)$. Weights are normalized within each anatomical axis; the eight weighted slice-age estimates form one sagittal, coronal or axial age prediction, and the final scan-level prediction is the equal-weight mean of the three axis estimates.

Training minimizes scan-level Huber loss \cite{huber1964} plus 0.25 times a Gaussian-weighted slice-level Huber loss, with a Huber transition parameter of 5 years. Optimization uses AdamW \cite{loshchilov2019}, an encoder learning rate of $10^{-5}$, a voting-head learning rate of $3\times10^{-4}$, weight decay $10^{-4}$, five warm-up epochs followed by cosine decay, age-bin sample weighting, gradient clipping at norm 5 and automatic mixed precision when supported. Training uses batch size two per GPU for up to 120 epochs, with validation-MAE early stopping with patience 15.

Held-out evaluation reports MAE, RMSE, $R^2$, Pearson correlation, Spearman correlation, median absolute error and brain-age delta. Confidence intervals for MAE, RMSE, $R^2$ and Pearson $r$ are estimated with 1,000 bootstrap resamples using seed 2027. Axis-specific, chronological-age-bin and sex-stratified metrics are reported descriptively from the same fixed held-out predictions.

\subsection*{Statistical evaluation}
For MNIST and CIFAR-10, top-1 confidence intervals are Wilson intervals \cite{wilson1927} and planned scratch-versus-transfer comparisons use exact McNemar tests \cite{mcnemar1947} with Holm multiplicity adjustment \cite{holm1979}. For chest radiography, confidence intervals for accuracy and macro-F1 are estimated from 300 bootstrap resamples \cite{efron1979}. For BrainAGE, confidence intervals for MAE, RMSE, $R^2$ and Pearson correlation are estimated from 1,000 bootstrap resamples \cite{efron1979} of the fixed 433-scan test set. Primary radiography metrics are accuracy, balanced accuracy and macro-F1; additional metrics include classwise precision and recall, macro and weighted AUROC/AUPRC, Matthews correlation coefficient, Cohen kappa, log loss, multiclass Brier score and 15-bin expected calibration error. Cross-dataset comparisons are descriptive because the datasets differ in sample structure, target semantics and evaluation protocol.

\section*{Ethics and data governance}
This study performed secondary analyses of existing image datasets and did not prospectively recruit participants or collect new human-subject data. The BrainAGE cohort uses source structural-neuroimaging data distributed by the originating repositories; use of those data remains subject to the governance, access and data-use terms of the respective repositories. The skin-disease benchmark uses the publicly accessible Kaggle dataset \cite{mishraSkinDisease}; the distributed files used in this study did not include patient or lesion identifiers, so patient-level grouping, linkage and re-identification were not performed. Restricted participant-level neuroimaging source or derived records are not redistributed by the authors.

\section*{Data availability}
MNIST, CIFAR-10, ImageNet-1K, the COVID-19 Radiography Database, the Skin Disease Dataset \cite{mishraSkinDisease}, and the publicly accessible or controlled-access neuroimaging source datasets underlying the BrainAGE cohort are available from their respective repositories subject to source-specific access, licensing and data-use conditions. The skin-disease source images are available through Kaggle. The BrainAGE analysis uses T1-weighted MRI from the same 2,851-scan source cohort described in Methods; restricted participant-level source or derived data must be obtained directly from the originating repositories under the applicable data-use agreements. ConnectomeX source code and author-generated manuscript-reproduction resources are proprietary and managed by the Columbia Technology Ventures Office of Intellectual Property. Data supporting the findings of this study are available from the corresponding author J.G. upon reasonable request where permitted by applicable third-party data-use agreements, repository terms and Columbia University intellectual-property requirements.

\section*{Code availability}
The ConnectomeX source code and associated manuscript-reproduction resources are proprietary and managed by the Columbia Technology Ventures Office of Intellectual Property and are not publicly released at the time of this preprint.

\section*{Acknowledgements}
Data collection and sharing for the Alzheimer's Disease Neuroimaging Initiative (ADNI) is funded by the National Institute on Aging (National Institutes of Health Grant U19AG024904). The grantee organization is the Northern California Institute for Research and Education. In the past, ADNI has also received funding from the National Institute of Biomedical Imaging and Bioengineering, the Canadian Institutes of Health Research, and private-sector contributions facilitated through the Foundation for the National Institutes of Health (FNIH), including AbbVie; Alzheimer's Association; Alzheimer's Drug Discovery Foundation; Araclon Biotech; BioClinica, Inc.; Biogen; Bristol-Myers Squibb Company; CereSpir, Inc.; Cogstate; Eisai Inc.; Elan Pharmaceuticals, Inc.; Eli Lilly and Company; EuroImmun; F. Hoffmann-La Roche Ltd and its affiliated company Genentech, Inc.; Fujirebio; GE Healthcare; IXICO Ltd.; Janssen Alzheimer Immunotherapy Research \& Development, LLC; Johnson \& Johnson Pharmaceutical Research \& Development LLC; Lumosity; Lundbeck; Merck \& Co., Inc.; Meso Scale Diagnostics, LLC; NeuroRx Research; Neurotrack Technologies; Novartis Pharmaceuticals Corporation; Pfizer Inc.; Piramal Imaging; Servier; Takeda Pharmaceutical Company; and Transition Therapeutics. The Canadian Institutes of Health Research supports ADNI clinical sites in Canada; the study is coordinated by the Alzheimer's Therapeutic Research Institute at the University of Southern California, and ADNI data are disseminated by the Laboratory for Neuro Imaging at the University of Southern California.

Data were provided in part by OASIS. The OASIS Cross-Sectional and Longitudinal projects were led by D. Marcus, R. Buckner, J. Csernansky and J. Morris (with A. Fotenos contributing to the Longitudinal study) and were supported by P50 AG05681, P01 AG03991, P01 AG026276, R01 AG021910, P20 MH071616 and U24 RR021382. Data were also provided in part by the Brain Genomics Superstruct Project of Harvard University and Massachusetts General Hospital (Principal Investigators Randy Buckner, Joshua Roffman and Jordan Smoller), with support from the Center for Brain Science Neuroinformatics Research Group, the Athinoula A. Martinos Center for Biomedical Imaging and the Center for Human Genetic Research. Twenty individual investigators at Harvard and Massachusetts General Hospital contributed data to the overall project.

Data used in preparation of this article were obtained from the Australian Imaging, Biomarkers and Lifestyle flagship study of ageing (AIBL), funded by the Commonwealth Scientific and Industrial Research Organisation (CSIRO) and made available through the ADNI database. AIBL researchers contributed data but did not participate in the ConnectomeX analysis or writing; the AIBL investigator group is listed at \href{https://aibl.csiro.au}{aibl.csiro.au}.

Data used in this study were also obtained from the Frontotemporal Lobar Degeneration Neuroimaging Initiative (FTLDNI/NIFD) database (\href{https://ida.loni.usc.edu/login.jsp?project=NIFD}{ida.loni.usc.edu}). FTLDNI data collection was supported by the National Institute on Aging (NIH Grant R01 AG032306), coordinated through the University of California, San Francisco Memory and Aging Center, and disseminated by the Laboratory for Neuro Imaging at the University of Southern California. The NIFD/FTLDNI investigators contributed to the design and implementation of the source initiative and/or provided data but did not participate in the ConnectomeX analysis or writing unless listed as authors. The FTLDNI investigator group associated with this source cohort includes Howard Rosen, Bradford C. Dickerson, Kimoko Domoto-Reilly, David Knopman, Bradley F. Boeve, Adam L. Boxer, John Kornak, Bruce L. Miller, William W. Seeley, Maria-Luisa Gorno-Tempini, Scott McGinnis and Maria Luisa Mandelli.

Data were provided in part by the Dallas Lifespan Brain Study (DLBS) \cite{park2025dlbs}. We thank the DLBS participants and investigators for making these data available. The DLBS was supported by the National Institute on Aging through grants 5R37AG-006265-27 and RC1AG036199. Data were also provided by the Southwest University Adult Lifespan Dataset (SALD) \cite{wei2018sald}, distributed through FCP/INDI. The SALD repository acknowledges support from the National Natural Science Foundation of China (31470981, 31571137 and 31500885), the National Outstanding Young People Plan, the Program for the Top Young Talents by Chongqing, Fundamental Research Funds for the Central Universities (SWU1509383, SWU1509451 and SWU1609177), the Natural Science Foundation of Chongqing (cstc2015jcyjA10106), the Fok Ying Tung Education Foundation (151023), the China Postdoctoral Science Foundation (2015M572423 and 2015M580767), the Chongqing Postdoctoral Science Foundation (Xm2015037 and Xm2016044), and Key Research for Humanities and Social Sciences of the Ministry of Education (14JJD880009). Data were additionally provided through the Consortium for Reliability and Reproducibility (CoRR) \cite{zuo2014corr}, hosted by the 1000 Functional Connectomes Project/International Data-sharing Initiative (FCP/INDI); we acknowledge the CoRR investigators, contributing sites and study participants who made the shared resource possible.

Data used in preparation of this article were obtained from the Parkinson's Progression Markers Initiative (PPMI) database (\href{https://www.ppmi-info.org/access-data-specimens/data}{ppmi-info.org/access-data-specimens/data}). PPMI is a public--private partnership funded by the Michael J. Fox Foundation for Parkinson's Research and its funding partners; current sponsor information is available at \href{https://www.ppmi-info.org/about-ppmi/who-we-are/study-sponsors}{ppmi-info.org}. Data were also obtained from the SchizConnect database (\href{http://schizconnect.org}{schizconnect.org}); investigators within SchizConnect contributed to the design and implementation of SchizConnect and/or provided data but did not participate in the ConnectomeX analysis or writing. SchizConnect data collection and sharing were supported by NIMH cooperative agreement 1U01MH097435. Data from the Southwest University Longitudinal Imaging Multimodal (SLIM) study were accessed through the International Data-sharing Initiative (INDI; \href{http://fcon_1000.projects.nitrc.org/}{fcon\_1000.projects.nitrc.org}). Data were additionally provided in part by IXI (\href{https://brain-development.org/ixi-dataset/}{brain-development.org/ixi-dataset}).

The authors acknowledge the use of OpenAI ChatGPT (GPT-5.6 and GPT-6) and OpenAI Codex for assistance with some of the code development and debugging, figure preparation, and manuscript proofreading. All AI-assisted outputs were reviewed and verified by the authors, who take responsibility for the scientific content.

\begingroup
\footnotesize
\setlength{\parskip}{0pt}
\bibliographystyle{unsrt}
\bibliography{references}
\endgroup
\clearpage
\section*{Figures}

\begin{figure}[!htbp]\centering
\includegraphics[width=0.88\textwidth]{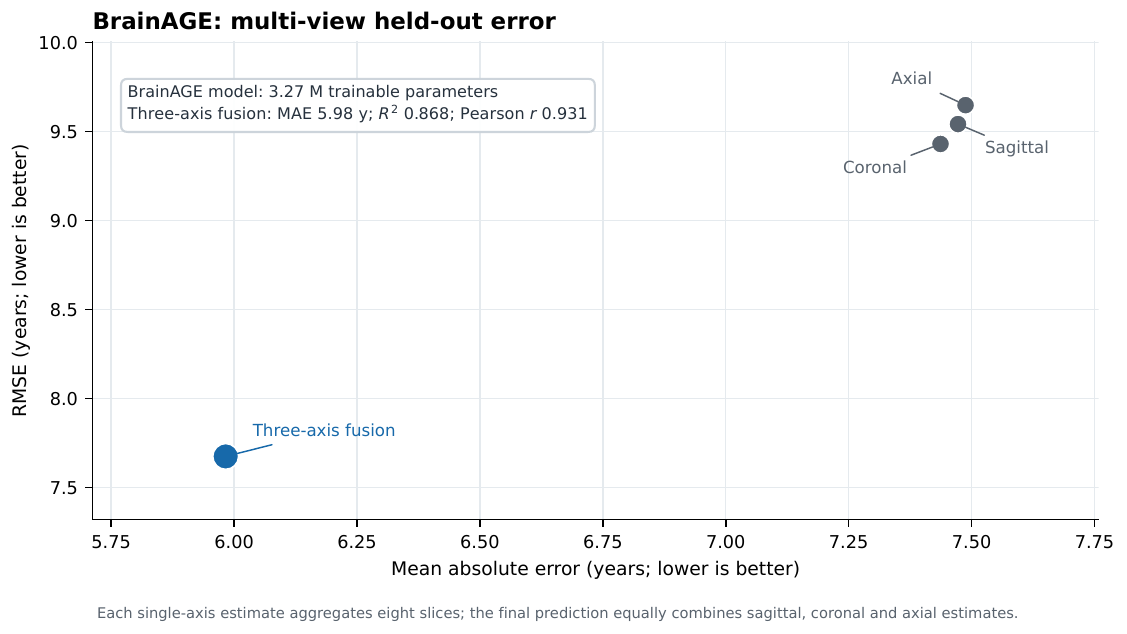}
\caption{BrainAGE multi-view aggregation overview. The complete BrainAGE model contains \BrainModelParamsM{} million trainable parameters. Axis-specific and three-axis fused predictions are positioned by held-out MAE and RMSE; the fused result is MAE \BrainMAE{} years, RMSE \BrainRMSE{} years, $R^2=\BrainRtwo$ and Pearson $r=\BrainPearson$ on \BrainTestN{} held-out scans.}\label{fig:brainage-overview}
\end{figure}

\begin{figure}[!htbp]\centering
\includegraphics[width=0.80\textwidth]{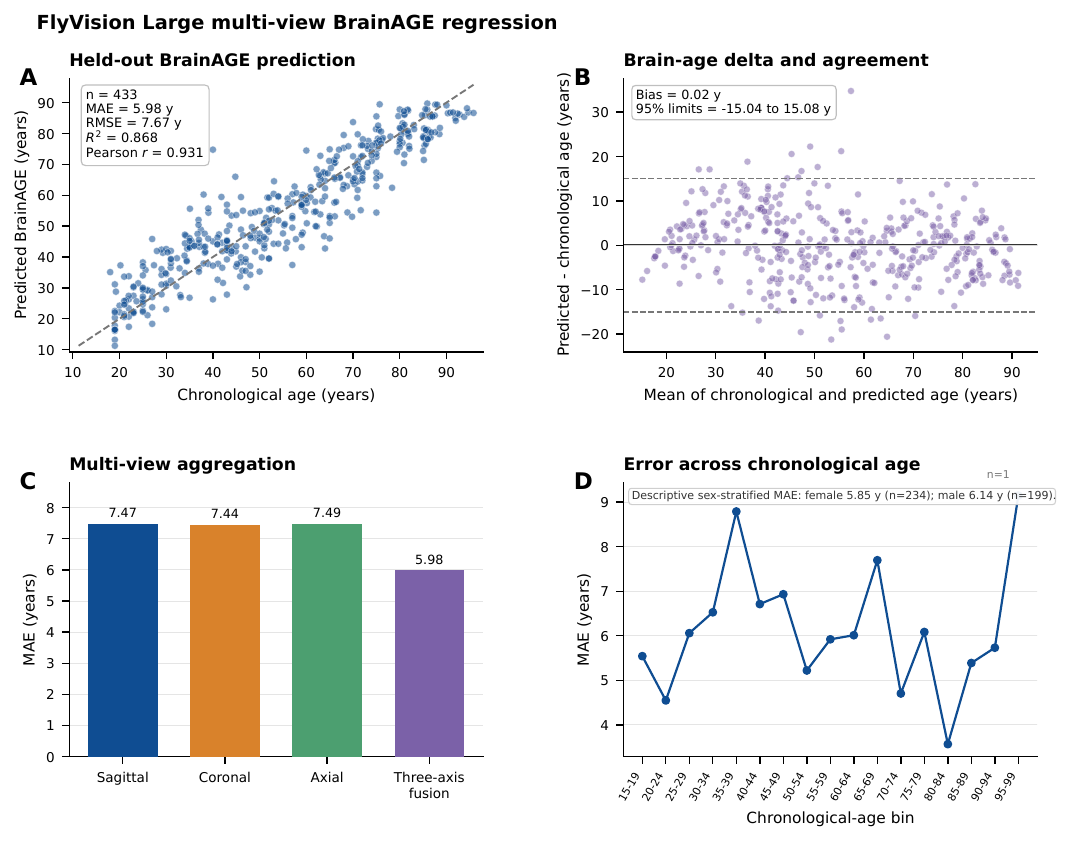}
\caption{Multi-view FlyVision Large BrainAGE regression. (A) Predicted versus chronological age for \BrainTestN{} held-out scans; the dashed diagonal denotes identity and the inset reports the primary regression metrics. (B) Bland--Altman representation of brain-age delta against the mean of predicted and chronological age. The solid horizontal line marks mean delta and dashed lines mark the 95\% limits of agreement. (C) MAE for sagittal, coronal and axial axis-specific estimates and for final three-axis fusion. Each axis estimate aggregates eight native-scale slices using the Gaussian positional prior and learned slice confidence. (D) MAE across five-year chronological-age bins; sex-stratified MAE is shown descriptively in the inset.}\label{fig:brainage}
\end{figure}

\clearpage
\section*{Tables}

\begin{table}[!htbp]\centering\caption{MNIST test performance.}\label{tab:mnist}\small
\begin{tabular}{lrrrr}\toprule Model & Initialization & Parameters & Accuracy (\%) & Macro-F1 (\%)\\\midrule
ResNet18 & ImageNet transfer & 11,181,642 & 99.38 & 99.38\\
FlyVision & Scratch & 80,608 & 99.34 & 99.33\\
ResNet18 & Scratch & 11,181,642 & 99.22 & 99.22\\
LeNet & Scratch & 44,426 & 98.60 & 98.59\\
MLP & Scratch & 109,386 & 97.50 & 97.48\\\bottomrule
\end{tabular}\end{table}

\begin{table}[!htbp]\centering\caption{CIFAR-10 test performance and transfer conditions.}\label{tab:cifar}\small
\begin{tabular}{llrrr}\toprule Model & Initialization & Parameters & Accuracy (\%) & Macro-F1 (\%)\\\midrule
ResNet18 & ImageNet direct & 11,181,642 & 85.84 & 85.81\\
ResNet18 & ImageNet$\rightarrow$MNIST & 11,181,642 & 84.04 & 84.10\\
ResNet18 & Scratch & 11,181,642 & 83.44 & 83.41\\
ResNet18 & MNIST transfer & 11,181,642 & 82.47 & 82.52\\
FlyVision & Scratch & 81,408 & 78.03 & 77.99\\
FlyVision & MNIST transfer & 81,408 & 77.62 & 77.55\\
LeNet & MNIST transfer & 62,006 & 72.71 & 72.68\\
LeNet & Scratch & 62,006 & 70.87 & 70.77\\
MLP & MNIST transfer & 402,250 & 53.70 & 53.12\\
MLP & Scratch & 402,250 & 52.60 & 51.70\\\bottomrule
\end{tabular}\end{table}

\begin{table}[!htbp]\centering\caption{ImageNet-1K validation performance at best-top-1 checkpoints. Top-5 is paired with the selected top-1 checkpoint. GFLOPs count forward Conv/Linear operations for 224$\times$224 inputs with two FLOPs per multiply-accumulate for the ImageNet execution graph (stride-2 stem; S1/S2/S3 strides 1/2/2 and recurrent-update counts 1/2/2); the later local/low-frequency model was not profiled under this convention.}\label{tab:imagenet}\scriptsize
\begin{tabular}{lrrrr}\toprule Model & Parameters (M) & GFLOPs & Top-1 (\%) & Top-5 (\%)\\\midrule
FlyVision Base, global & 1.836 & 6.632 & 60.79 & 82.33\\
FlyVision Base, local-k7 & 1.836 & -- & 60.25 & 81.93\\
FlyVision Large, global & 3.737 & 14.803 & 66.25 & 86.07\\
FlyVision Large, local-k7 + low-frequency & 3.744 & -- & 66.53 & 86.31\\
ResNet18, scratch & 11.690 & 3.628 & 69.25 & 88.57\\\bottomrule
\end{tabular}\end{table}

\begin{table}[!htbp]\centering\caption{Skin-disease test performance on the identical 1,538-image held-out set. Macro-AUROC is the unweighted mean of 22 one-versus-rest class AUROCs.}\label{tab:skin}\scriptsize
\begin{tabular}{lrrrrr}\toprule Model & Parameters (M) & Accuracy (\%) & Macro-F1 (\%) & Balanced accuracy (\%) & Macro-AUROC (\%)\\\midrule
FlyVision Base, global & 1.334 & 60.01 & 55.47 & 55.24 & 94.43\\
FlyVision Base, local-k7 & 1.334 & 58.45 & 54.34 & 53.88 & 94.56\\
FlyVision Base, local-k7 + LF & 1.339 & 62.42 & 58.67 & 58.06 & 95.23\\
FlyVision Large, global & 2.985 & 63.39 & 59.94 & 59.80 & 95.59\\
FlyVision Large, local-k7 + LF & 2.992 & 63.78 & 60.23 & 58.28 & 95.28\\
ResNet18 & 11.188 & 66.19 & 61.96 & 62.27 & 95.96\\
\bottomrule\end{tabular}\end{table}

\begin{table}[!htbp]\centering\caption{Chest-radiography test performance and model efficiency. All models were evaluated on the same 3,175-image held-out test split. GFLOPs are measured from the executed radiography forward graphs for 224$\times$224 inputs with two FLOPs per multiply-accumulate. The checkpoint-loaded FlyVision radiography graph uses stage strides 2/2/2 and one recurrent update per stage, whereas the ImageNet graph uses stage strides 1/2/2 and recurrent-update counts 1/2/2; macro-AUROC uses one-versus-rest aggregation.}\label{tab:cxr}\small
\begin{tabular}{lrrrrr}\toprule Model & Parameters (M) & GFLOPs & Accuracy (\%) & Macro-F1 (\%) & Macro-AUROC (\%)\\\midrule
FlyVision Large (pretrained) & 2.97 & 2.79 & 92.76 & 92.85 & 98.68\\
FlyVision Base (pretrained) & 1.33 & 1.28 & 92.60 & 92.75 & 98.68\\
FlyVision Base (scratch) & 1.33 & 1.28 & 92.19 & 92.09 & 98.67\\
ResNet18 & 11.18 & 3.63 & 91.56 & 91.98 & 98.01\\
LeNet & 0.46 & 0.14 & 89.80 & 89.83 & 97.68\\
MLP & 19.28 & 0.04 & 56.91 & 50.98 & 79.88\\\bottomrule
\end{tabular}\end{table}

\begin{table}[!htbp]\centering\caption{Hierarchical chest-radiography performance.}\label{tab:hier}\small
\begin{tabular}{lrr}\toprule System & Accuracy (\%) & Macro-F1 (\%)\\\midrule
Flat FlyVision four-class & 92.60 & 92.75\\
Two-stage hierarchical & 91.53 & 91.68\\
Multi-head hierarchical & 91.75 & 92.17\\\midrule
Disease-only stage-2 subtype & 95.02 & 95.58\\
Multi-head subtype & 95.57 & 96.09\\\bottomrule
\end{tabular}\end{table}

\begin{table}[!htbp]\centering\caption{BrainAGE held-out regression values. The three-axis fusion row reports 95\% bootstrap confidence intervals for the primary metrics; axis-specific rows report fixed held-out errors.}\label{tab:brainage}\small
\begin{tabular}{lrrrrr}\toprule
Estimate & n & MAE (years) & RMSE (years) & $R^2$ & Pearson $r$\\\midrule
Three-axis fusion & \BrainTestN & \BrainMAE{} [\BrainMAECILow--\BrainMAECIHigh] & \BrainRMSE{} [\BrainRMSECILow--\BrainRMSECIHigh] & \BrainRtwo{} [\BrainRtwoCILow--\BrainRtwoCIHigh] & \BrainPearson{} [\BrainPearsonCILow--\BrainPearsonCIHigh]\\
Sagittal axis & \BrainTestN & \BrainSagMAE & \BrainSagRMSE & -- & --\\
Coronal axis & \BrainTestN & \BrainCorMAE & \BrainCorRMSE & -- & --\\
Axial axis & \BrainTestN & \BrainAxMAE & \BrainAxRMSE & -- & --\\
\bottomrule
\end{tabular}
\end{table}

\end{document}